\documentclass[10pt,journal,compsoc]{IEEEtran}

\usepackage{times}
\usepackage{graphicx}
\usepackage{amsmath}
\usepackage{amssymb}
\usepackage{tabularx}
\usepackage{multirow}
\usepackage{float}%
\usepackage[export]{adjustbox} %
\usepackage[para]{threeparttable} %
\usepackage{enumitem} %
\usepackage{booktabs, siunitx, dcolumn}
\usepackage{algorithm}     %
\usepackage{algorithmic}     %
\usepackage{array} %
\usepackage{url} %
\usepackage{fixltx2e} %
\usepackage{color}
\usepackage{dashrule} %
\ifCLASSOPTIONcompsoc
  \usepackage[nocompress]{cite}
\else
  \usepackage{cite}
\fi

\ifCLASSOPTIONcompsoc
  \usepackage[caption=false,font=footnotesize,labelfont=sf,textfont=sf]{subfig}
\else
  \usepackage[caption=false,font=footnotesize]{subfig}
\fi

\begin{document}

\title{Proposal Flow: Semantic Correspondences from Object Proposals}

\author{Bumsub~Ham,~\IEEEmembership{Member,~IEEE,}
        Minsu~Cho,
		Cordelia~Schmid,~\IEEEmembership{Fellow,~IEEE}
        and~Jean~Ponce,~\IEEEmembership{Fellow,~IEEE}%
\IEEEcompsocitemizethanks{\IEEEcompsocthanksitem Bumsub Ham is with the School of Electrical and Electronic Engineering, Yonsei University, Seoul, Korea. E-mail: mimo@yonsei.ac.kr.
\IEEEcompsocthanksitem Minsu Cho is with the Department of Computer Science and Engineering, POSTECH, Pohang, Korea. E-mail: mscho@postech.ac.kr.
\IEEEcompsocthanksitem Cordelia Schmid is with Thoth project-team, Inria Grenoble Rh\^one-Alpes, Laboratoire Jean Kuntzmann, France. E-mail: cordelia.schmid@inria.fr.
\IEEEcompsocthanksitem Jean Ponce is with {\'E}cole Normale Sup{\'e}rieure / PSL Research University and WILLOW project-team (CNRS/ENS/INRIA UMR 8548), Paris, France. E-mail: jean.ponce@ens.fr.}%
}

\markboth{SUBMISSION TO IEEE TRANSACTIONS ON PATTERN ANALYSIS AND MACHINE INTELLIGENCE, 2017}%
{SUBMISSION TO IEEE TRANSACTIONS ON PATTERN ANALYSIS AND MACHINE INTELLIGENCE, 2017}

\IEEEtitleabstractindextext{
\begin{abstract}
Finding image correspondences remains a challenging problem in the presence of intra-class variations and large changes in scene layout.~Semantic flow methods are designed to handle images depicting different instances of the same object or scene category. We introduce a novel approach to semantic flow, dubbed proposal flow, that establishes reliable correspondences using object proposals. Unlike prevailing semantic flow approaches that operate on pixels or regularly sampled local regions, proposal flow benefits from the characteristics of modern object proposals, that exhibit high repeatability at multiple scales, and can take advantage of both local and geometric consistency constraints among proposals. We also show that the corresponding sparse proposal flow can effectively be transformed into a conventional dense flow field. We introduce two new challenging datasets that can be used to evaluate both general semantic flow techniques and region-based approaches such as proposal flow. We use these benchmarks to compare different matching algorithms, object proposals, and region features within proposal flow, to the state of the art in semantic flow. This comparison, along with experiments on standard datasets, demonstrates that proposal flow significantly outperforms existing semantic flow methods in various settings.
\end{abstract}

}

\maketitle

\IEEEdisplaynontitleabstractindextext
\IEEEpeerreviewmaketitle

\IEEEraisesectionheading{\section{Introduction}}

\IEEEPARstart{C}{lassical} approaches to finding correspondences across images are designed to handle scenes that contain the same objects with moderate view point variations in applications such as stereo matching~\cite{okutomi1993multiple,rhemann2011fast}, optical flow~\cite{horn1993determining,weinzaepfel2015deepmatching,weinzaepfel2013deepflow}, and wide-baseline matching~\cite{matas2004robust,yang2014daisy}. {\em Semantic flow} methods, such as SIFT Flow~\cite{liu2011sift} for example, on the other hand, are designed to handle a much higher degree of variability in appearance and scene layout, typical of images depicting different instances of the same object or scene category.~They have proven useful for many tasks such as object recognition, cosegmentation, image registration, semantic segmentation, and image editing and synthesis~\cite{hacohen2011non,kim2013deformable,liu2011sift,yang2014daisy,zhou2015flowweb, duchenne2011graph, taniai2016joint}. 
In this context, however, appearance and shape variations may confuse similarity measures for local region matching, and prohibit the use of strong geometric constraints (e.g., epipolar geometry, limited disparity range). Existing approaches to semantic flow are thus easily distracted by scene elements specific to individual objects and image-specific details (e.g., background, texture, occlusion, clutter). This is the motivation for our work, where we use reliable and robust region correspondences to focus on regions containing prominent objects and scene elements rather than clutter and distracting details.

\begin{figure}[t]
\centering
\includegraphics[width=\linewidth]{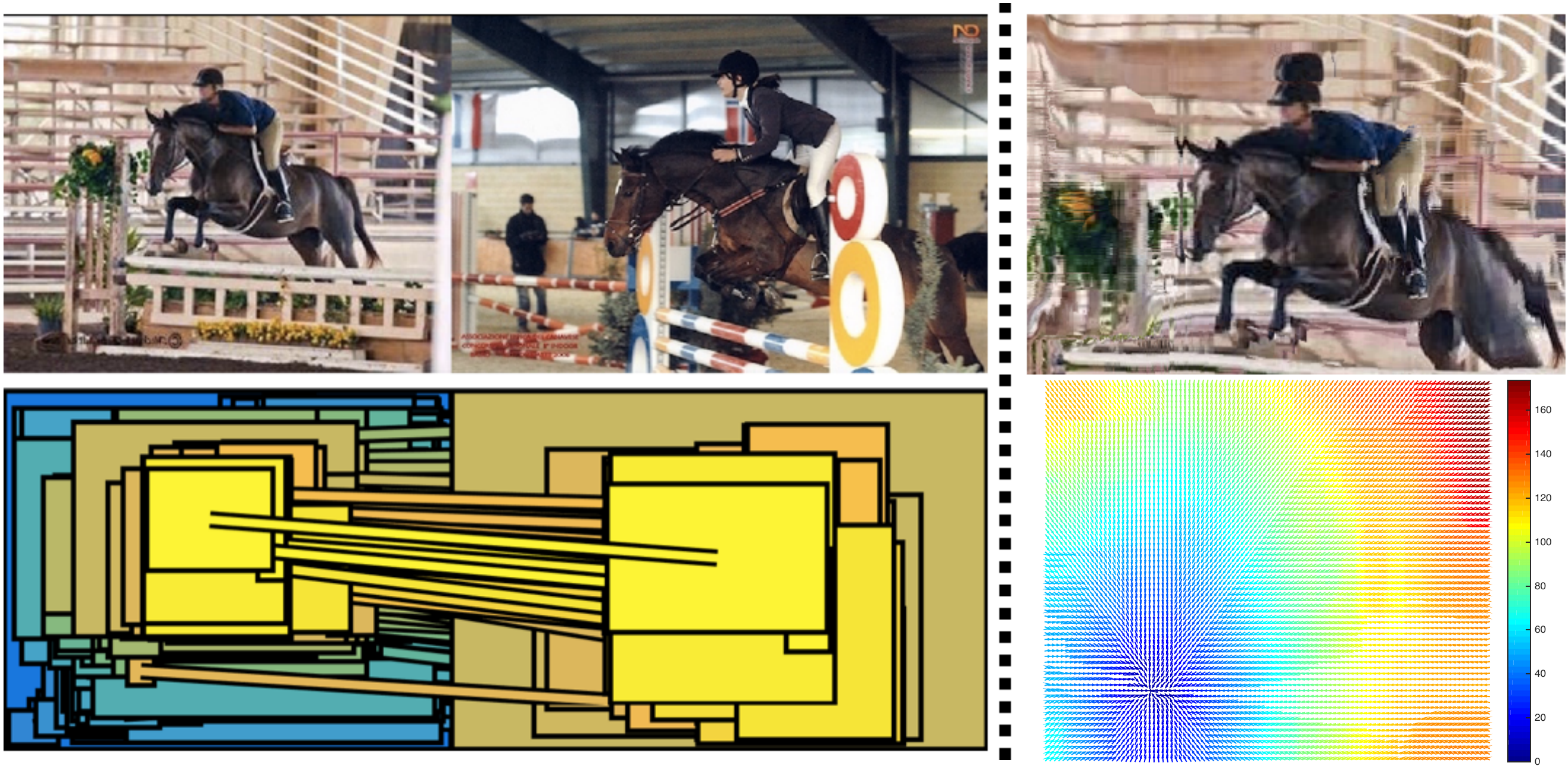}
\vspace{-0.1cm}
\begin{minipage}{0.625\linewidth}
\centering
\small{(a) Region-based semantic flow.}
\end{minipage}
\vspace{-0.1cm}
\begin{minipage}{0.365\linewidth}
\centering
\small{(b) Dense flow field.}
\end{minipage}
\caption{Proposal flow generates a reliable and robust semantic flow between similar images using local and geometric consistency constraints among object proposals, and it can be transformed into a dense flow field. Using object proposals for semantic flow enables focusing on regions containing prominent objects and scene elements rather than clutter and distracting details. (a) Region-based semantic flow between source (left) and target (right) images.~(b) Dense flow field (bottom) and image warping using the flow field (top). \textbf{(Best viewed in color.)}}
\label{fig:teaser}	
\end{figure}

Concretely, we introduce an approach to pairwise semantic flow
computation, called {\em proposal flow}, that establishes region
correspondences using object proposals and their geometric relations
(Fig.~\ref{fig:teaser}). Unlike previous semantic flow
algorithms~\cite{BristowVL15,hacohen2011non,hassner2012sifts,hur2015generalized,kim2013deformable,liu2011sift,qiu2014scale,tau2014dense,trulls2013dense,yang2014daisy,zhou2015flowweb,choy2016universal,taniai2016joint, zhou2016learning},
that use regular grid structures for local region generation and
matching, we leverage a large number of multi-scale object
proposals~\cite{arbelaez2014multiscale,hosang2015what,manen2013prime,uijlings2013selective,zitnick2014edge},
as now widely used to significantly reduce the search space or false alarms,~e.g, for object
detection~\cite{girshickICCV15fastrcnn,kaiming14ECCV} and tracking tasks~\cite{zhu2016beyond}. Using object proposals for semantic flow has the following advantages: First, we can use diverse spatial supports for prominent objets and parts, and focus on these elements rather than clutter and distracting scene components. Second, we can use geometric relations between objects and parts, which prevents confusing objects with visually similar regions or parts, but quite different geometric configurations. Third, as in the case of object detection, we can reduce the search space for correspondences, scaling well with the size of the image collection.  Accordingly, the proposed approach establishes region correspondences between object proposals by exploiting their visual features and geometric relations in an efficient manner, and generates
a region-based semantic flow composed of object proposal matches.~We show that this region-based proposal flow can be effectively transformed into a
conventional dense flow field. We also introduce new datasets and evaluation metrics that can be used to evaluate both general semantic flow techniques and
region-based approaches such as proposal flow. These datasets consist of images containing more clutter and intra-class variation, and are much more challenging than existing ones for semantic flow evaluation. We use these benchmarks to compare different matching algorithms, object proposals, and region features within proposal flow, to the state of the art in semantic flow. This comparison, along with experiments on standard datasets,
demonstrates that proposal flow significantly outperforms existing semantic flow methods~(including a learning-based approach) in various settings.

{\textbf{Contributions.}} The main contributions of this paper can be summarized as follows:
\begin{itemize}[leftmargin=*]
	\item We introduce the proposal flow approach to establishing robust region
correspondences between related, but not identical scenes using object
proposals (Section~\ref{sec:pf}).
\item We introduce benchmark datasets and evaluation metrics for semantic flow that can be
used to evaluate both general semantic flow algorithms and region
matching methods (Section~\ref{sec:benchmark}).
\item We demonstrate the
advantage of proposal flow over state-of-the-art semantic
flow methods through extensive experimental evaluations (Section~\ref{sec:exp}).
\end{itemize}
A preliminary version of this work appeared in~\cite{ham2016proposal}.~Besides a more detailed presentation and discussion of the most recent related works, this version adds (1)~an in-depth presentation of proposal flow;~(2)~a more challenging benchmark based on the PASCAL 2011 keypoint dataset~\cite{bourdev2009poselets};~(3)~a verification of quality of ground-truth correspondence generation for our datasets;~(4) an extensive experimental evaluation including a performance analysis with varying the number of proposals and an analysis of runtime, and a comparison of proposal flow with recently introduced state-of-the-art methods and datasets. To encourage comparison and future work, our datasets and code are available online: http://www.di.ens.fr/willow/research/proposalflow.

\section{Related work} 
Correspondence problems involve a broad range of topics beyond the scope of this paper. Here we briefly describe the context of our approach, and only review representative works pertinent for ours. 
\subsection{Semantic flow}
\textbf{Pairwise correspondence.}
Classical approaches to stereo matching and optical flow estimate dense correspondences between pairs of nearby images of the same scene~\cite{horn1993determining,matas2004robust,okutomi1993multiple}. 
While advances in invariant feature detection and description have revolutionized object recognition and reconstruction in the past 15 years, research on image matching and alignment between images have long been dominated by instance matching with the same scene and objects~\cite{forsyth2011modern}. Unlike these, several recent approaches to semantic flow focus on handling images containing different scenes and objects.~Graph-based matching algorithms~\cite{cho2012progressive,duchenne2011graph} attempt to find category-level feature matches by leveraging a flexible graph representation of images, but they commonly handle sparsely sampled or detected features due to their computational complexity. Inspired by classic optical flow algorithms, Liu~\emph{et al}. pioneered the idea of dense correspondences across different scenes, and proposed the SIFT Flow~\cite{liu2011sift} algorithm that uses a multi-resolution image pyramid together with a hierarchical optimization technique for efficiency. Kim~\emph{et al}.~\cite{kim2013deformable} extende the approach by inducing a multi-scale regularization with a hierarchically connected pyramid of grid graphs. Long~\emph{et al}.~\cite{long2014convnets} investigate the effect of pretrained ConvNet features on the SIFT Flow algorithm, and Bristow~\emph{et al}.~\cite{BristowVL15} propose an exemplar-LDA approach that improves the performance of semantic flow. More recently, Taniai~\emph{et al}.~\cite{taniai2016joint} have shown that the approach to jointly recovering cosegmentation and dense correspondence outperforms state-of-the-art methods designed specifically for either cosegmentation or correspondence estimation. Zhou~\emph{et al}.~\cite{zhou2016learning} propose a learning-based method that leverages a 3D model. This approach uses cycle consistency to link the correspondence between real images and rendered views. Choy~\emph{et al}.~\cite{choy2016universal} propose to use a fully convolutional architecture, along with a correspondence contrastive loss, allowing faster training by effective reuse of computations. While archiving state-of-the-art performance, these learning-based approaches require a large number of annotated images~\cite{choy2016universal} or 3D models~\cite{zhou2016learning} to train the corresponding deep model, and do not consider geometric consistency among correspondences. 

Despite differences in graph construction, optimization, and similarity computation, existing semantic flow approaches share grid-based regular sampling and spatial regularization: The appearance similarity is defined at each region or pixel on (a pyramid of)  regular grids, and spatial regularization is imposed between neighboring regions in the pyramid models~\cite{kim2013deformable,liu2011sift,long2014convnets,taniai2016joint}. In contrast, our work builds on generic object proposals with diverse spatial supports~\cite{arbelaez2014multiscale,hosang2015what,manen2013prime,uijlings2013selective,zitnick2014edge}, and uses an irregular form of spatial regularization based on co-occurrence and overlap of the proposals. We show that the use of local regularization with object proposals yields substantial gains in generic region matching and semantic flow, in particular when handling images with significant clutter, intra-class variations and scaling changes, establishing a new state of the art on the task.

\textbf{Multi-image correspondence.}
Besides these pairwise matching methods, recent works have tried to solve a correspondence problem as a joint image-set alignment. Collection Flow~\cite{kemelmacher2012collection} uses an optical flow algorithm that aligns each image to its low-rank projection onto a sub-space capturing the common appearance of the image collection. FlowWeb~\cite{zhou2015flowweb} first builds a fully-connected graph with each image as a node, and each edge as flow field between a pair of images, and then establishes globally-consistent correspondences using cycle consistency among all edges. This approach gives state-of-the-art performance, but requires a large number of images for each object category, and the matching results are largely dependent on the initialization quality. Zhou \emph{et al}.~\cite{zhou2015multi} also use cycle consistency between sparse features to solve a graph matching problem posed as a low-rank matrix recovery. Carreira~\emph{et al}.~\cite{carreira2015virtual} leverage keypoint annotations to estimate dense correspondences across images with similar viewpoint, and use these pairwise matching results to align a query image to all the other images to perform single-view 3D reconstruction. 

While improving over pairwise correspondence results at the expense of runtime, these multi-image methods all use a pairwise method to find initial matches before refining them,~(e.g., with cycle consistency~\cite{zhou2015multi}). Our correspondence method outperforms current pairwise methods, and its output could be used as a good initialization for multi-image methods.

\subsection{Object proposals and object-centric representations}
Object proposals~\cite{arbelaez2014multiscale,hosang2015what,manen2013prime,uijlings2013selective,zitnick2014edge} have originally been developed  for object detection, where they are used to reduce the search space as well as false alarms. They are now an important component in many state-of-the-art detection pipelines~\cite{girshickICCV15fastrcnn,kaiming14ECCV} and other computer vision applications, including object tracking~\cite{zhu2016beyond}, action recognition~\cite{gkioxari2015contextual}, weakly supervised localization~\cite{cinbis2014multi}, and semantic segmentation~\cite{dai2015boxsup}. Despite their success for object detection and segmentation, object proposals have seldom been used in matching tasks~\cite{cho2015unsupervised,jiang2015matching}. In particular, while Cho \emph{et al}.~\cite{cho2015unsupervised} have shown that object proposals are useful for region matching due to their high repeatability on salient part regions, the use of object proposals has never been thoroughly investigated in semantic flow computation. The approach proposed in this paper is a first step in this direction, and we explore how the choice of object proposals, matching algorithms, and features affects matching robustness and accuracy. 

Recently, object-centric representation has been used to estimate optical flow. In~\cite{bai2016deep}, potentially moving vehicles are first segmented from the background, and the flow is estimated individually for every object and the background. Similarly, Sevilla-Lara~\emph{et al}.~\cite{sevilla2016optical} use semantic segmentation to break the image into regions, and compute optical flow differently in different regions, depending on the the semantic class label. The main intuition behind these works is that focusing on regions containing prominent regions, e.g., objects, can help estimate the optical flow field effectively. Proposal flow shares similar idea, but it is designed for semantic flow computation and leverages the geometric relations between objects and parts as well. We show that object proposals are well suited to semantic flow computation, and further using their geometric relations boosts the matching accuracy. 

\section{Proposal flow}\label{sec:pf}

Proposal flow can use any type of object
proposals~\cite{arbelaez2014multiscale,hosang2015what,manen2013prime,ren15fasterrcnn,uijlings2013selective,zitnick2014edge}
as candidate regions for matching a pair of images of related scenes. In
this section, we introduce a probabilistic model for region matching~(Section~\ref{subsec:bayesian}),
and describe three matching strategies including two baselines and a 
new one using local regularization~(Section~\ref{subsec:matching}). We then describe our approach
to generating a dense flow field from the region matches~(Section~\ref{subsec:flowfield}).

\begin{figure*}
\centering
	\subfloat[Input images.]{
	\includegraphics[width=0.326\textwidth, frame]{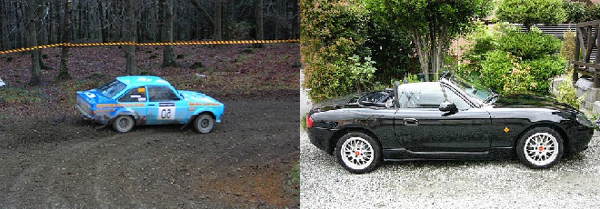}}
	\subfloat[Object proposals~\cite{uijlings2013selective}.]{
	\includegraphics[width=0.326\textwidth, frame]{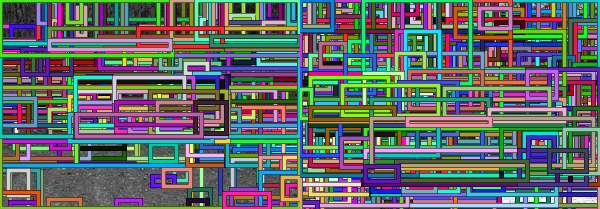}}
	\subfloat[Object proposals near the front wheel.]{
	\includegraphics[width=0.326\textwidth, frame]{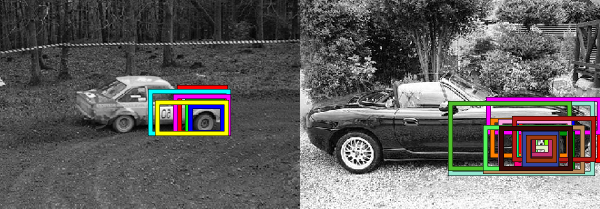}}

	\subfloat[NAM.]{
	\includegraphics[width=0.326\textwidth, frame]{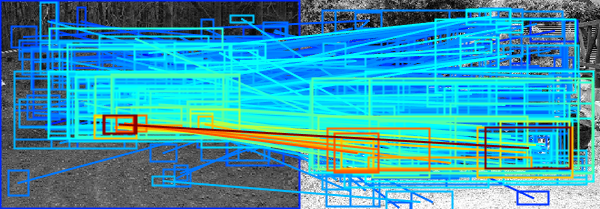}}
	\subfloat[PHM~\cite{cho2015unsupervised}.]{
	\includegraphics[width=0.326\textwidth, frame]{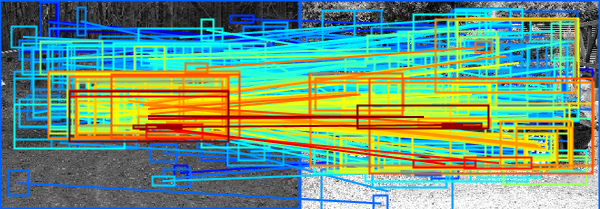}}
	\subfloat[LOM.]{
	\includegraphics[width=0.326\textwidth, frame]{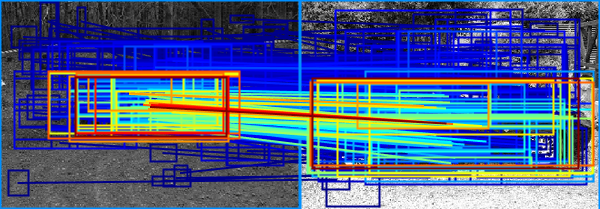}}
	\vfill
\caption{ \textbf{Top:} (a-b) A pair of images and their object
  proposals~\cite{uijlings2013selective}.  (c) Multi-scale object
  proposals contain the same object or parts, but they are not
  perfectly repeatable across different images. \textbf{Bottom:} In
  contrast to NAM (d), PHM~\cite{cho2015unsupervised} (e) and LOM (f)
  both exploit geometric consistency, which regularizes proposal
  flow. In particular, LOM imposes local smoothness on offsets between
  neighboring regions, avoiding the problem of using a global
  consensus on the offset in PHM~\cite{cho2015unsupervised}. The
  matching score is color-coded for each match (red: high, blue:
  low). The HOG descriptor~\cite{dalal2005histograms} is used for
  appearance matching in this example. \textbf{(Best viewed in color.)}}
\label{fig:compare_match}	
\end{figure*}

\subsection{A Bayesian model for region matching}\label{subsec:bayesian}

Let us suppose that two sets of object proposals $\mathcal{R}$ and $\mathcal{R}'$ have been extracted from images $\mathcal{I}$ and $\mathcal{I}'$ (Fig.~\ref{fig:compare_match}(a-b)). A proposal $r$ in $\mathcal{R}$ is an image region $r=(f,s)$ with appearance feature $f$ and spatial support $s$.
The appearance feature represents a visual descriptor for the region (e.g., SPM~\cite{lazebnik2006beyond}, HOG~\cite{dalal2005histograms}, ConvNet~\cite{krizhevsky2012imagenet}), and the spatial support describes the set of all pixel positions in the region~(a rectangular box in this work). Given the data $\mathcal{D} = (\mathcal{R}, \mathcal{R}')$, we wish to estimate a posterior probability of the event $r \mapsto r'$ meaning that proposal $r$ in $\mathcal{R}$ matches proposal $r'$ in $\mathcal{R}'$:  
\begin{equation}
\label{eq:similairty}
	p(r \mapsto r' \mid \mathcal{D}) = p(f \mapsto f' ) p(s \mapsto s' \mid \mathcal{D}), 
\end{equation}
where we decouple appearance and geometry, and further assume that appearance matching is independent of the data $\mathcal{D}$.
In practice, the appearance term $p(f \mapsto f')$ is simply computed from a similarity between feature descriptors $f$ and $f'$, and the geometric consistency term $p(s \mapsto s' \mid \mathcal{D})$ is evaluated by comparing the spatial supports $s$ and $s'$ in the context of the given data $\mathcal{D}$, as described in the next section. We set the posterior probability $p(r \mapsto r' \mid \mathcal{D})$ as a matching score and assign the best match $\phi(r)$ for each proposal in $\mathcal{R}$:   
\begin{equation}
\label{eq:max}
	\phi(r)=\mathop {\text{argmax} }_{r'\in \mathcal{R^\prime} } p(r \mapsto r' \mid \mathcal{D}).  
\end{equation}
Using a slight abuse of notation, if $(f',s')=\phi(f,s)$, we will write $f'=\phi(f)$ and $s'=\phi(s)$.

\subsection{Geometric matching strategies}\label{subsec:matching}

We now introduce three matching strategies, using different geometric consistency terms $p(s \mapsto s' \mid \mathcal{D})$.

\subsubsection{Naive appearance matching (NAM)}
A straightforward way of matching regions is to use a uniform distribution for the geometric consistency term $p(s \mapsto s' \mid \mathcal{D})$ so that    
\begin{equation}
\label{eq:naive}
	p(r \mapsto r' \mid \mathcal{D}) \propto p(f \mapsto f').
\end{equation}
NAM considers appearance only, and does not reflect any geometric relationship among regions (Fig. \ref{fig:compare_match}(d)).

\subsubsection{Probabilistic Hough matching (PHM)}
The matching algorithm in~\cite{cho2015unsupervised} can be expressed in our model as follows. First, a three-dimensional location vector (position plus scale) is extracted from the spatial support $s$ of each proposal $r$. We denote it by a function $\gamma$. An offset space $\mathcal{X}$ is defined as a feasible set of offset vectors between $\gamma(s)$ and $\gamma(s')$: $\mathcal{X}=\{\gamma(s) - \gamma(s') \mid r \in \mathcal{R}, r' \in \mathcal{R}' \}$. The geometric consistency term $p(s \mapsto s' \mid \mathcal{D})$ is then defined as 
\begin{equation}
	p(s \mapsto s' \mid \mathcal{D}) = \sum_{x \in \mathcal{X}} p(s \mapsto s' \mid x) p(x \mid \mathcal{D}), 
\end{equation}
which assumes that the probability $p(s \mapsto s' \mid x)$ that two boxes $s$ and $s'$ match given the offset $x$ is independent of the rest of the data and can be modeled by a Gaussian kernel in the three-dimensional offset space. Given this model, PHM replaces $p(x \mid \mathcal{D})$ with a generalized Hough transform score:\vspace{-0.1cm}   
\begin{equation}
	h(x \mid \mathcal{D} ) = \sum_{ (r,r') \in \mathcal{D}} p(f \mapsto f') p(s \mapsto s' \mid x),
\end{equation}
which aggregates individual votes for the offset $x$, from \emph{all} possible matches in $\mathcal{D}=\mathcal{R} \times \mathcal{R}'$. Hough voting imposes a spatial regularizer on matching by taking into account a \emph{global} consensus on the corresponding offset~\cite{leibe2008robust,maji2009object}. However, it often suffers from background clutter that distracts the global voting process (Fig. \ref{fig:compare_match}(e)).

\subsubsection{Local offset matching (LOM)}
Here we propose a new method to overcome this drawback of PHM~\cite{cho2015unsupervised} and obtain more reliable correspondences. Object proposals often contain a large number of distracting outlier regions from background clutter, and are not perfectly repeatable even for corresponding object or parts across different images (Fig. \ref{fig:compare_match}(c)). The global Hough voting in PHM has difficulties with such outlier regions. In contrast, we optimize a translation and scale offset for each proposal by exploiting only neighboring proposals. That is, instead of averaging $p(s \mapsto s' | x)$ over all feasible offsets $\mathcal{X}$ in PHM, we use one reliable offset optimized for each proposal. This local approach substantially alleviates the effect of outlier regions in matching as will be demonstrated by our experiment results.

The main issue is how to estimate a reliable offset for each proposal $r$ in a robust manner without any information about objects and their locations. One way would be to find the region corresponding to $r$ through a multi-scale sliding window search in $\mathcal{I}'$ as in object detection~\cite{felzenszwalb2008discriminatively}, but this is expensive. Instead, we assume that nearby regions have similar offsets. For each region $r$, we first define its neighborhood $\mathcal{N}(r)$ as the set of regions with overlapping spatial support:  
\begin{equation}
	\mathcal{N}(r)=\{\hat r \mid s \cap \hat s  \neq \emptyset, \hat r \in \mathcal{R} \}.   
\end{equation}
Using an initial correspondence $\psi(r)$, determined by the best match according to appearance, each neighboring region $\hat r$ is assigned its own offset, and all of them form a set of neighbor offsets:   
\begin{equation}
	\mathcal{X}(r) =\{ \gamma(\hat s) - \gamma(\psi(\hat s)) \mid \hat r \in \mathcal{N}(r) \}.  
\end{equation} 
 From this set of neighbor offsets, we estimate a local offset $x^*_r$ for the region $r$ by the geometric median~\cite{lopuhaa1991breakdown}\footnote{We found that the centroid and mode of the offset vectors in three-dimensional offset space show worse performance than the geometric median. This is because the neighboring regions may include clutter. Clutter causes incorrect neighbor offsets, but the geometric median is robust to outliers~\cite{fletcher2008robust}, providing a reliable local offset.}:  
\begin{equation}
	x^*_r =\mathop {\text{argmin}}_{x \in \mathbb{R}^3} \sum_{y \in \mathcal{X}(r)} \left\| x-y \right\|_2,  
\end{equation}
which can be computed using Weiszfeld's algorithm~\cite{chandrasekaran1989open} with a form of iteratively re-weighted least squares. In other words, the local offset $x^*_r$ for the region $r$ is estimated by regression using its local neighboring offsets $\mathcal{X}(r)$. Based on the local offset $x^*_r$ optimized for each region, we define the geometric consistency function:   
\begin{equation}
	g(s \mapsto s' | \mathcal{D} ) = p(s \mapsto s' | x^*_r)\sum_{\hat r \in \mathcal{N}(r)} p(\hat f \mapsto \psi(\hat f)),  
\end{equation}
which can be interpreted as the fact that the region $r$ in $\mathcal{R}$ is likely to match $r'$ in $\mathcal{R}^\prime$ where its offset $\gamma(s) - \gamma(s')$ is close to the local offset $x^*_r$, and the region $r$ has many neighboring matches with a high appearance fidelity. By using $g(s \mapsto s' | \mathcal{D} )$ as a proxy for $p(s \mapsto s' | \mathcal{D} )$, LOM imposes local smoothness on offsets between neighboring regions. This geometric consistency function effectively suppresses matches between clutter regions, while favoring matches between regions that contain objects rather than object parts (Fig.~\ref{fig:compare_match}(f)). 

\begin{figure}
\centering
	\subfloat[Anchor match and pixel correspondence.]{
	\includegraphics[width=0.475\textwidth, frame]{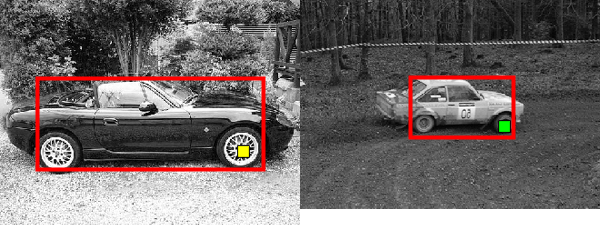}}

	\subfloat[Match visualization.]{
	\includegraphics[width=0.235\textwidth, height=0.17\textwidth, frame]{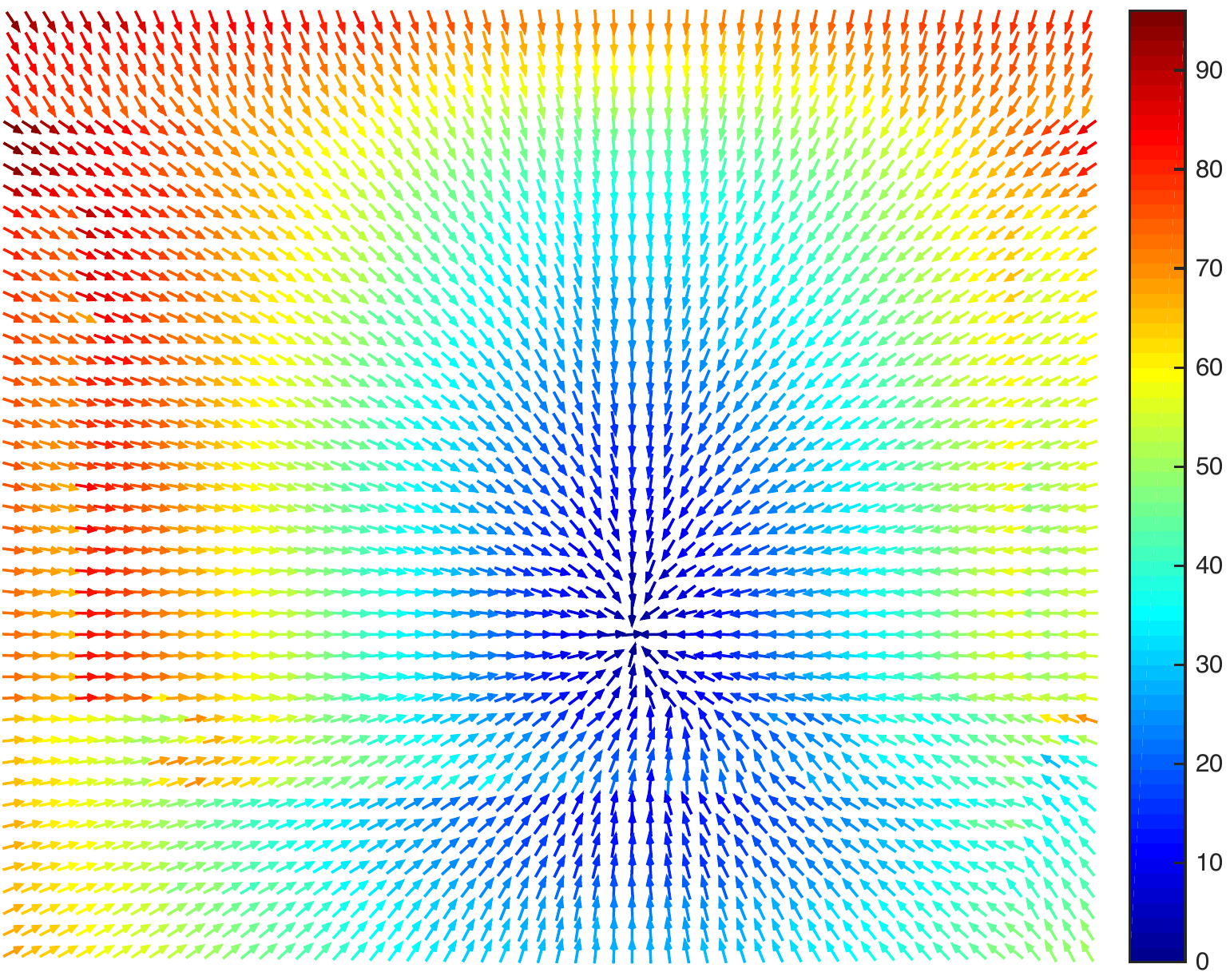}}
	\subfloat[Warped image.]{
	\includegraphics[width=0.235\textwidth, height=0.17\textwidth, frame]{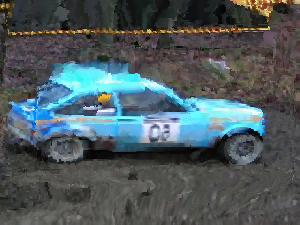}}
	\vfill
\caption{Flow field generation. (a) For each pixel (yellow point), its anchor match (red boxes) is determined. The correspondence (green point) is computed by the transformed coordinate with respect to the position and size of the anchor match. (b) Based on the flow field, (c) the right image is warped to the left image. The warped object shows visually similar shape to the one in the left image. The LOM method is used for region matching with the object proposals~\cite{manen2013prime} and the HOG descriptor~\cite{dalal2005histograms}. \textbf{(Best viewed in color.)}}
\label{fig:anchor_match}	
\end{figure}

\begin{figure*}
\centering
\subfloat[Keypoints and object bounding boxes.]{
\includegraphics[width=0.326\textwidth, frame]{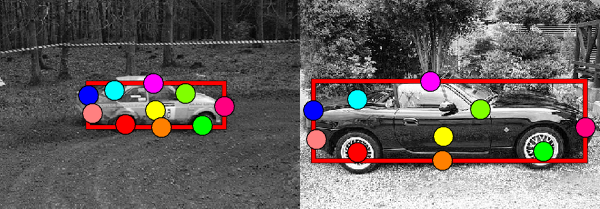}}
\subfloat[Warping.]{
\includegraphics[width=0.163\textwidth, frame]{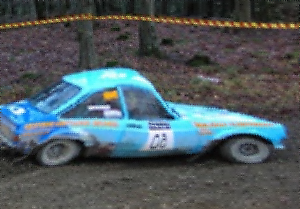}}
\subfloat[$\mathcal{R}_s$.]{
\includegraphics[width=0.163\textwidth, frame]{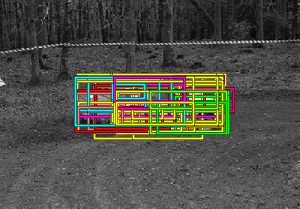}}
\subfloat[Ground-truth correspondence]{
\includegraphics[width=0.326\textwidth, frame]{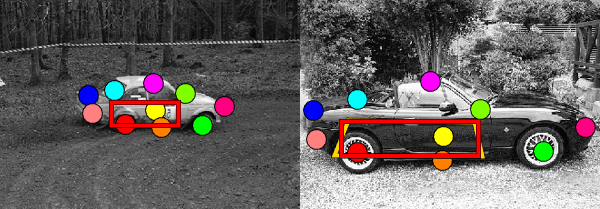}}

\subfloat[NAM.]{
\includegraphics[width=0.326\textwidth, frame]{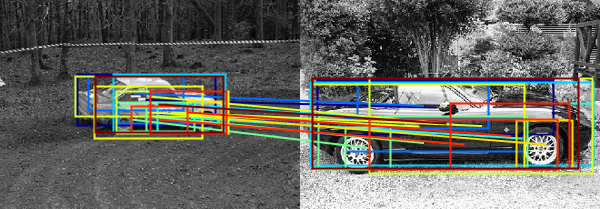}}
\subfloat[PHM~\cite{cho2015unsupervised}.]{
\includegraphics[width=0.326\textwidth, frame]{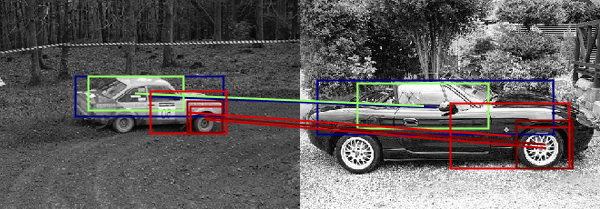}}
\subfloat[LOM.]{
\includegraphics[width=0.326\textwidth, frame]{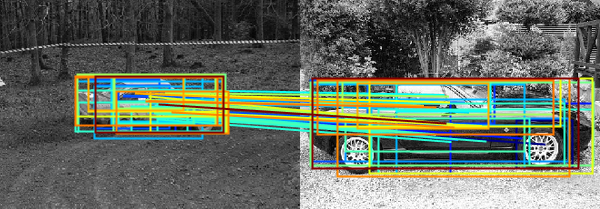}}
\vfill

\caption{\textbf{Top:} Generating ground-truth regions and evaluating correct
  matches. (a) Using keypoint annotations, dense correspondences
  between images are established using TPS warping~\cite{bookstein1989principal, donato2002approximate}. (b)
  Based on the dense correspondences, all pixels in the left image are
  warped to the right image, showing that the correspondences align two images well. (c) We assume that true matches exist
  only between the regions near the object bounding box, and thus an
  evaluation is done with the regions in this subset of object
  proposals. (d) For each object proposal (red box in the left image),
  its ground truth is generated automatically by the dense
  correspondences: We fit a tight rectangle (red box in the right
  image) of the region formed by the warped object proposal (yellow
  box in the right image) and use it as a ground-truth correspondence. \textbf{Bottom:} Examples of correct matches: The numbers of correct matches are 16, 5,
  and 38 for NAM (e), PHM~\cite{cho2015unsupervised} (f), and LOM (g),
  respectively. Matches with an IoU score greater than 0.5 are
  considered as correct in this example. \textbf{(Best viewed in color.)}}
\label{fig:gt_generation}
\end{figure*}

\subsection{Flow field generation}\label{subsec:flowfield}

Proposal flow gives a set of region correspondences between images that can easily be transformed into a conventional dense flow field. 
Let $p$ denote a pixel in the image $\mathcal{I}$ (yellow point in Fig.~\ref{fig:anchor_match}(a)). For each pixel $p$, its neighborhood is defined as the region in which it lies, i.e., $\mathcal{N}(p)=\{ r\in \mathcal{R}:p \in r \}$. We define an anchor match $(r^*,\phi(r^*))$ as the region correspondence that has the highest matching score among neighboring regions (red boxes in Fig.~\ref{fig:anchor_match}(a)) where   
\begin{equation}\label{eq:anchor_match}
	r^*=\mathop {\text{argmax} }\limits_{r \in \mathcal{N}(p)} p(r \mapsto \phi(r) \mid \mathcal{D}).  
\end{equation}
Note that the anchor match contains information on translation and scale changes between objects or part regions. Using the geometric relationships between the pixel $p$ and its anchor match $(r^*,\phi(r^*))$, a correspondence $p'$ in the image $\mathcal{I}^\prime$ (green point in Fig. \ref{fig:anchor_match}(a)) is obtained by linear interpolation, i.e., computed by the transformed coordinate with respect to the position and size of the anchor match.

The matching score for each correspondence $p$ is set to the value of its anchor match $(r^*,\phi(r^*))$. When the pixels $p$ and $q$ in the image $\mathcal{I}$ are matched to the same pixel $p'$ in the image $\mathcal{I}'$, we select the match with the highest matching score and delete the other one. Finally, joint image filtering~\cite{ham2015robust} is applied under the guidance of the image $\mathcal{I}$ to interpolate the flow field in places without correspondences. Figure \ref{fig:anchor_match}(b-c) shows examples of the estimated flow field and corresponding warping result between two images: Using the dense flow field, we warp all pixels in the right image to the left image. Our approach using the anchor match aligns semantic object parts well while handling translation and scale changes between objects.

\section{Datasets for semantic flow evaluation}\label{sec:benchmark} 

Current research on semantic flow lacks appropriate benchmarks with
dense ground-truth correspondences. Conventional optical flow
benchmarks (e.g., Middlebury~\cite{baker2011database} and
MPI-Sintel~\cite{butler2012naturalistic}) do not feature within-class
variations, and ground truth for generic semantic flow is difficult to
capture due to its intrinsically semantic nature, manual annotation
being extremely labor intensive and somewhat subjective. Existing
approaches are thus usually evaluated only with sparse ground truth or in an
indirect manner (e.g. mask transfer
accuracy)~\cite{BristowVL15,kim2013deformable,liu2011sift,qiu2014scale,tau2014dense,zhou2015flowweb}.
Such benchmarks only evaluate a small number of matches, that occur at
ground-truth keypoints or around mask boundaries in a point-wise manner. 
To address this issue, we introduce in this section two new datasets for semantic flow, dubbed PF-WILLOW and PF-PASCAL (PF for proposal flow), built using ground-truth object bounding boxes and keypoint annotations, (Fig. \ref{fig:gt_generation}(a)), and propose new evaluation metrics for region-based semantic flow methods. Note that while designed for
region-based methods, our benchmark can be used to evaluate any semantic
flow technique. As will be seen in our experiments, it provides a
reasonable (if approximate) ground truth for dense correspondences
across similar scenes without an extremely expensive annotation
campaign. Comparative evaluations
on this dataset have also proven to be good predictors for performance on other
tasks and datasets, further justifying the use of our benchmark. 

Taniai~\emph{et al}. have recently introduced a benchmark dataset for semantic flow evaluation~\cite{taniai2016joint}. It provides 400 image pairs of 7 object categories, corresponding ground-truth cosegmentation masks, and flow maps that are obtained by natural neighbor interpolation~\cite{sibson1981brief} on sparse keypoint matches. In contrast, our datasets use over 2200+ image pairs of up to 20 categories. It is split into two subsets: The first subset features 900 image pairs of 4 object categories, further split into 10 sub-categories according to the viewpoint and background clutter, in order to evaluate the different factors of variation for matching accuracy. The second subset consists of 1300+ image pairs of 20 image categories. In the following, we present our ground-truth
generation process in Section~\ref{subsec:gt}, evaluation criteria in Section~\ref{subsec:metric}, and datasets in Section~\ref{subsec:dataset}. %

\subsection{Ground-truth correspondence generation}\label{subsec:gt}

Let us assume two sets of keypoint annotations at positions $k_i$ and $k'_i$ in $\mathcal{I}$ and $\mathcal{I}'$, respectively, with $i=1,\ldots,m$. Assuming the objects present in the images and their parts may undergo shape deformation, we use thin plate splines (TPS)~\cite{bookstein1989principal, donato2002approximate} to interpolate sparse keypoints~(Fig. \ref{fig:gt_generation}(b)). Concretely, the ground truth is approximated from sparse correspondences using TPS warping. 
For each region or proposal, its ground-truth match is generated as follows. We assume that each image has a single object and true matches only exist between a subset of regions, i.e., regions around object bounding boxes (Fig.~\ref{fig:gt_generation}(c)): $\mathcal{R}_s=\left\{r \mid {\left|b \cap r\right|} \mathbin{/} {\left| r \right|} \geq 0.75, r\in\mathcal{R} \right\}$ where $b$ denotes an object bounding box in the image $\mathcal{I}$, and $|r|$ indicates the area of the region $r$. %
For each region $r \in \mathcal{R}_s$ (e.g., red box in Fig. \ref{fig:gt_generation}(d)~left), the four vertices of the rectangle are warped to the corresponding ones in the image $\mathcal{I}'$ by the TPS mapping function (e.g., yellow box in Fig. \ref{fig:gt_generation}(d)~right). The region formed by the warped points is a correspondence of region $r$. 
We fit a tight rectangle for this region and set it as a ground-truth correspondence for the region $r$ (e.g., red box in Fig. \ref{fig:gt_generation}(d)~right). 

Note that WarpNet~\cite{kanazawa2016warpnet} also uses TPS to generate ground-truth correspondences, but it does not consider intra-class variation. In particular, WarpNet constructs a pose graph using a fine-grained dataset~(e.g., the CUB-200-2111~\cite{wah2011caltech} of bird categories), computes a set of TPS functions using silhouettes of image pairs that are closest on the graph, and finally transforms each image by sampling from this set of TPS warps. In contrast to this, we directly use TPS to estimate a warping function using ground-truth keypoint annotations.

\subsection{Evaluation criteria}\label{subsec:metric}

We introduce two evaluation metrics for region matching performance in terms of {\rm matching precision} and {\rm match retrieval accuracy}. These metrics build on the intersection over union (IoU) score between the region $r$'s correspondence ${\phi(r)}$ and its ground truth $r^\star$: 
\begin{equation} 
	\text{IoU}({\phi(r)},r^\star) = {|{\phi(r)} \cap r^\star|} \mathbin{/} { |{\phi(r)} \cup r^\star|}. 
\end{equation}
For region matching precision, we propose the probability of correct region (PCR) metric where the region $r$ is correctly matched to its ground truth $r^\star$ if $1-\text{IoU}({\phi(r)},r^\star) < \tau$ (e.g., Fig.~\ref{fig:benchmark}(a) top), where $\tau$ is an IoU threshold. Note that this region-based metric is based on a conventional point-based metric, the probability of correct keypoint (PCK)~\cite{yang2013articulated}. In the case of pixel-based flow, PCK can be adopted instead. We measure the PCR metric while varying the IoU threshold $\tau$ from 0 to 1. For match retrieval accuracy, we propose the average IoU of $k$-best matches (dubbed mIoU@$k$) according to the matching score (e.g., Fig. \ref{fig:benchmark}(a) bottom). We measure the mIoU@$k$ metric while increasing the number of top matches $k$. These two metrics exhibit two important characteristics of matching: PCR reveals the accuracy of overall assignment, and mIoU@$k$ shows the reliability of matching scores that is crucial in match selection. 

\subsection{Dataset construction}\label{subsec:dataset}

We construct two benchmark datasets for semantic flow evaluation: The PF-WILLOW and PF-PSCAL datasets. The original images and keypoint annotations are taken from existing datasets~\cite{bourdev2009poselets, cho2013learning}.

\textbf{PF-WILLOW.} To generate the PF-WILLOW dataset, we start from the benchmark for sparse
matching of Cho {\em et al.}~\cite{cho2013learning}, which consists of 5
object classes (Face, Car, Motorbike, Duck, WineBottle) with 10
keypoint annotations for each image. Note that these images contain
more clutter and intra-class variation than existing datasets~\cite{kim2013deformable,qiu2014scale,zhou2015flowweb} for semantic flow evaluation, which include mainly images with tightly cropped objects or similar background. We
exclude the face class where the number of generated object proposals
is not sufficient to evaluate matching accuracy. The other classes are
split into sub-classes\footnote{They are car (S), (G), (M), duck (S),
  motorbike (S), (G), (M), wine bottle (w/o C), (w/ C), (M), where (S)
  and (G) denote side and general viewpoints, respectively. (C) stands
  for background clutter, and (M) denotes mixed viewpoints (side +
  general) for car and motorbike classes and a combination of images
  in wine bottle (w/o C + w/ C) for the wine bottle class.} according
to viewpoint or background clutter. We obtain a total of 10
sub-classes. Given these images and regions, we generate ground-truth
data between all possible image pairs within each sub-class. The dataset
  has 10 images for each sub-class, thus 100 images and 900 image pairs in total. 

\textbf{PF-PASCAL.} For the PF-PASCAL dataset, we use PASCAL 2011 keypoint annotations~\cite{bourdev2009poselets} for 20 object categories. We select meaningful image pairs for each category that contain a single object with similar poses, resulting in 1351 image pairs in total. The number of image pairs in the dataset varies from 6 for the sheep class to 140 for the bus class, and 67 on average, and each image pair contains from 4 to 17 keypoints and 7.95 keypoints on average.  This dataset is more challenging than PF-WILLOW and other existing datasets for semantic flow evaluation. %

\section{Experiments}\label{sec:exp}
In this section we present a detailed analysis and evaluation of our proposal flow approach.

\subsection{Experimental details}

{\textbf{Object proposals.}}
We evaluate four state-of-the-art object proposal methods:~EdgeBox~(EB)~\cite{zitnick2014edge},~multi-scale combinatorial grouping~(MCG)~\cite{arbelaez2014multiscale},~selective search~(SS)~\cite{uijlings2013selective},~and randomized prim~(RP)~\cite{manen2013prime}.~In addition, we consider three baseline proposals~\cite{hosang2015what}:~Uniform sampling~(US), Gaussian sampling~(GS), and sliding window~(SW)~(See~\cite{hosang2015what} for a discussion). We use publicly available codes for all proposal methods. 

For fair comparison, we use 1,000 proposals for all the methods in all experiments, unless otherwise specified. To control the number of proposals, we use the proposal score: Albeit not all having explicit control over the number of proposals, EB, MCG, and SS provides proposal scores, so we use the top $k$ proposals. For RP, which lacks any control over the number of proposals, we randomly select the proposals. For US, GS, and SW, we can control the number of proposals explicitly~\cite{hosang2015what}.

\begin{figure*}
\centering
	\subfloat{
	\includegraphics[width=0.32\textwidth]{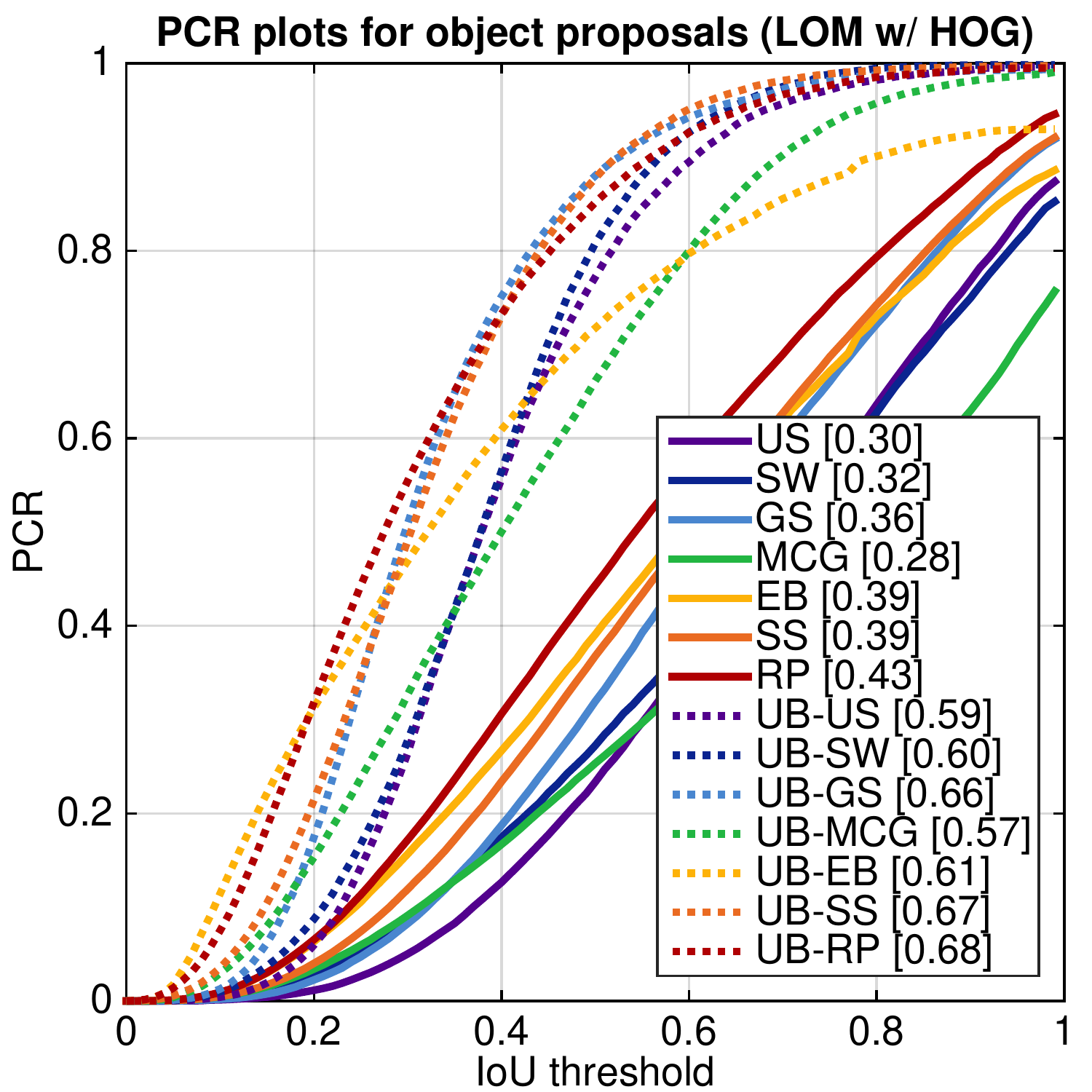}
	\includegraphics[width=0.32\textwidth]{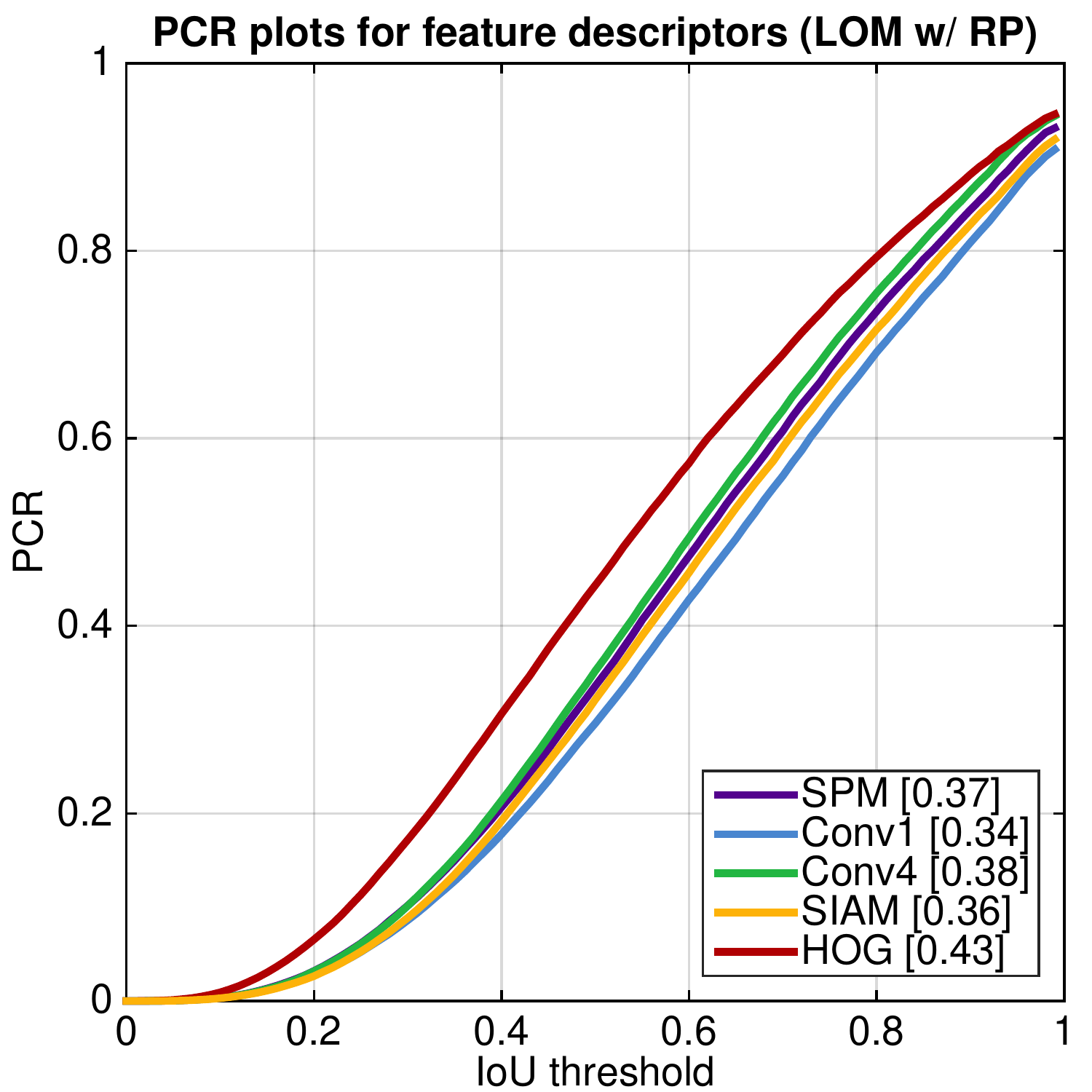}
	\includegraphics[width=0.32\textwidth]{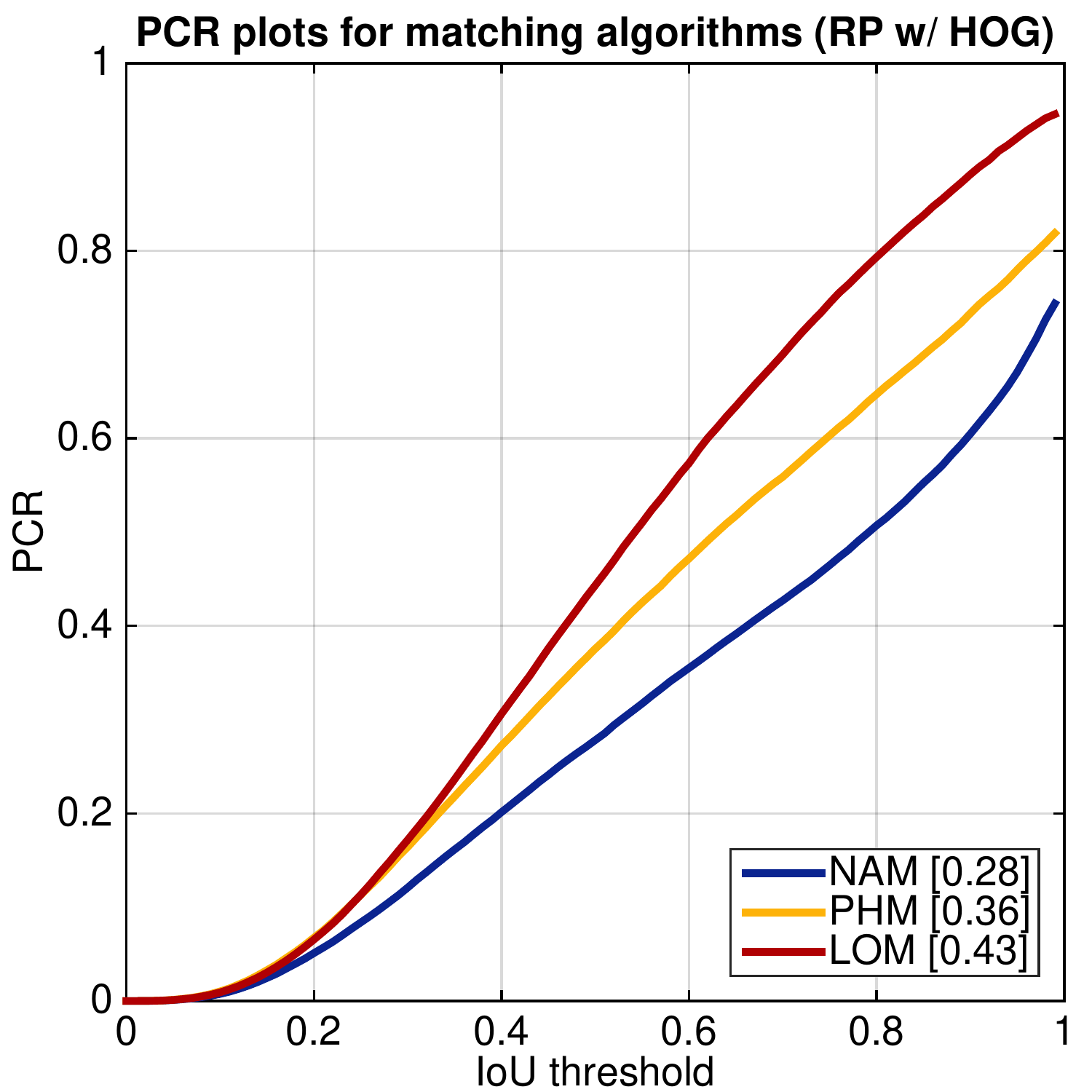}}

	\setcounter{subfigure}{0}
	\subfloat[Comparison of object proposals.]{
	\includegraphics[width=0.32\textwidth]{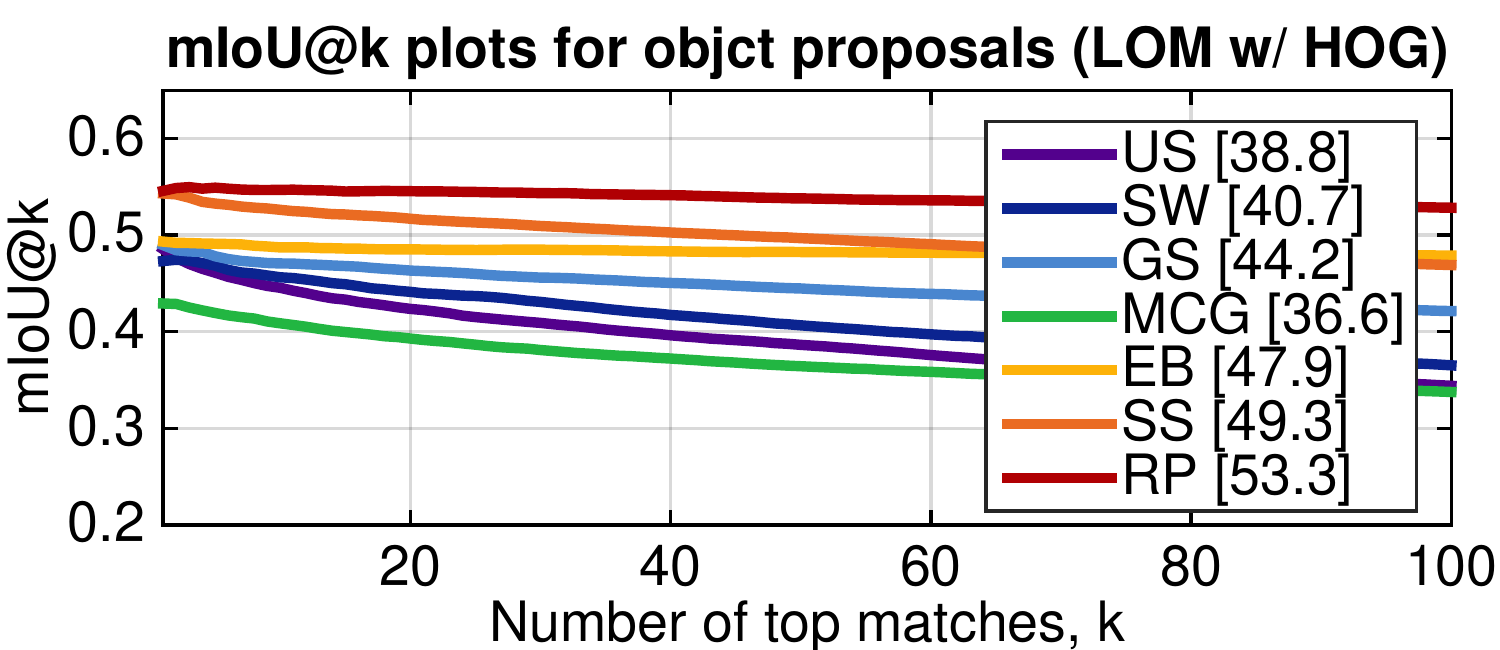}}
	\subfloat[Comparison of feature descriptors.]{
	\includegraphics[width=0.32\textwidth]{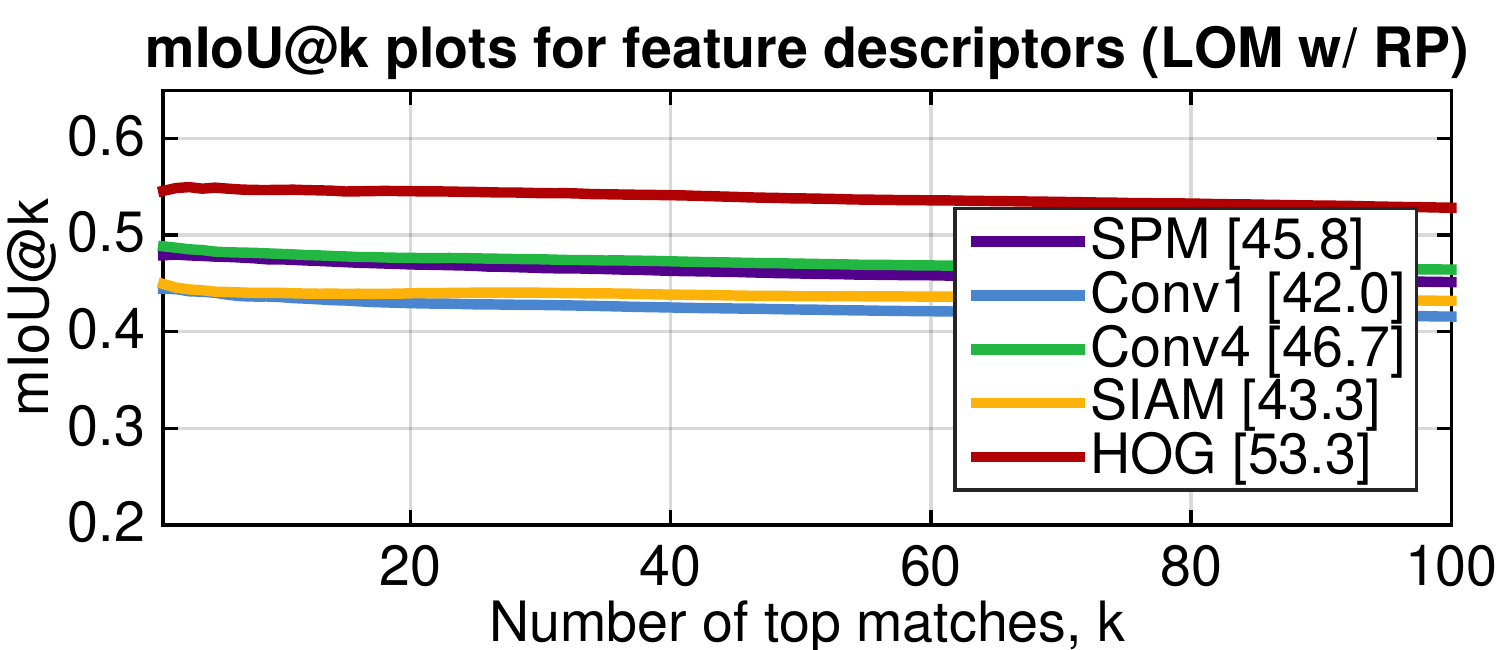}}	
	\subfloat[Comparison of matching algorithms.]{
	\includegraphics[width=0.32\textwidth]{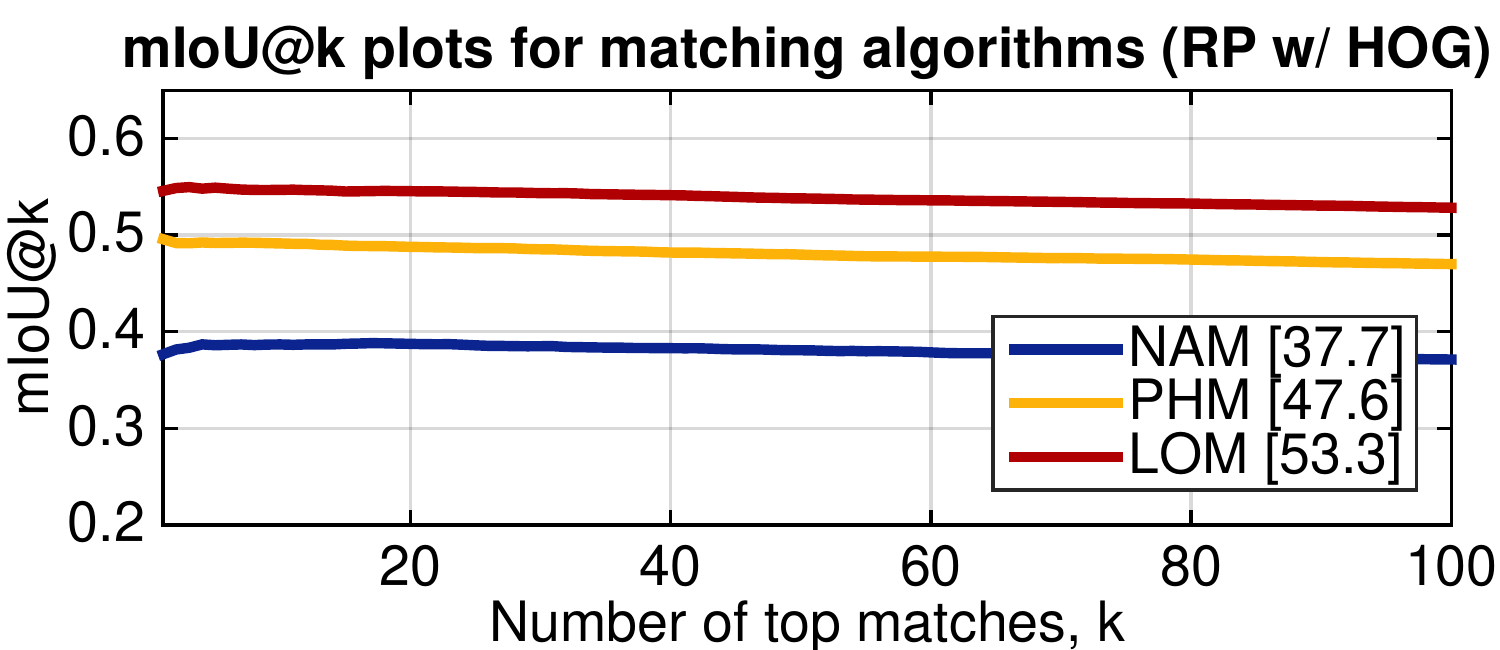}}	

\caption{PF-PASCAL benchmark evaluation on region matching precision (top, PCR plots) and match retrieval accuracy (bottom, mIoU@$k$ plots): (a) Evaluation for LOM with HOG~\cite{dalal2005histograms}, (b) evaluation for LOM with RP~\cite{manen2013prime}, and (c) evaluation for RP with HOG~\cite{dalal2005histograms}. The AuC is shown in the legend. \textbf{(Best viewed in color.)}}
\label{fig:benchmark}	
\end{figure*}

{\textbf{Feature descriptors and similarity.}}
We evaluate four popular feature descriptors: two engineered ones (SPM~\cite{lazebnik2006beyond} and HOG~\cite{dalal2005histograms}) and two learning-based ones (ConvNet~\cite{krizhevsky2012imagenet} and SIAM~\cite{simo2015discriminative}). For SPM, dense SIFT features~\cite{lowe2004distinctive} are extracted every 4 pixels and each descriptor is quantized into a 1,000 word codebook~\cite{tang2014co}. For each region, a spatial pyramid pooling~\cite{lazebnik2006beyond} is used with $1\times1$ and $3\times3$ pooling regions. We compute the similarity between SPM descriptors by the $\chi^2$ kernel. HOG features are extracted with $8\times8$ cells and 31 orientations, then whitened. For ConvNet features, we use each output of the 5 convolutional layers in AlexNet~\cite{krizhevsky2012imagenet}, which is pre-trained on the ImageNet dataset~\cite{deng2009imagenet}. For HOG and ConvNet, the dot product is used as a similarity metric\footnote{We also tried the $\chi^2$ kernel to compute the similarity between HOG or ConvNet features, and found that using the dot product gives better matching accuracy.}. For SIAM, we use the author-provided model trained using a Siamese network on a subset of Liberty, Yosemite, and Notre Dame images of the multi-view stereo correspondence (MVS) dataset~\cite{brown2011discriminative}. Following~\cite{simo2015discriminative}, we compute the similarity between SIAM descriptors by the $l_2$ distance.

\subsection{Proposal flow components}\label{subsec:ex-region-matching}

We use the PF benchmarks in this section to compare three variants of
proposal flow using different matching algorithms (NAM, PHM, LOM), combined with various object
proposals~\cite{arbelaez2014multiscale,hosang2015what,manen2013prime,uijlings2013selective,zitnick2014edge},
and features~\cite{dalal2005histograms,krizhevsky2012imagenet,lazebnik2006beyond, simo2015discriminative}.

\begin{figure*}
\centering
	\subfloat{
	\includegraphics[width=0.95\textwidth]{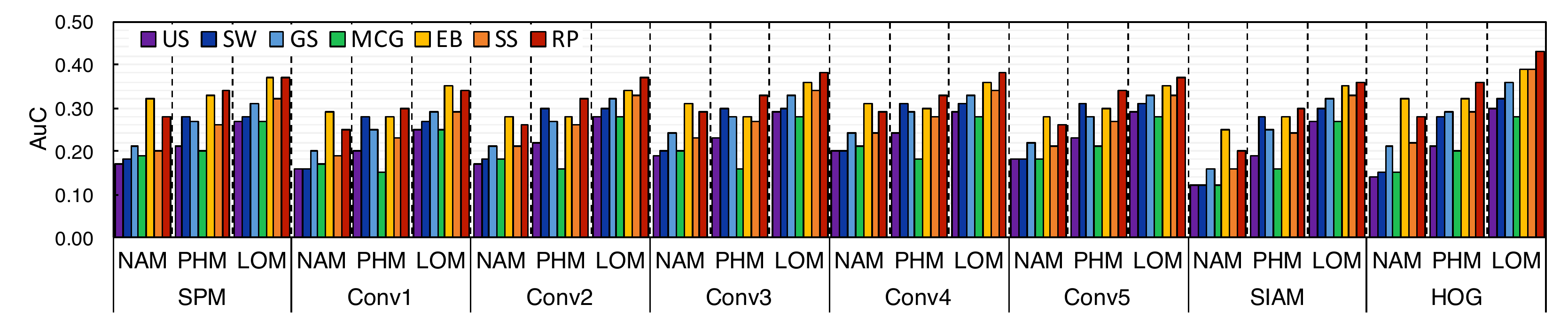}}
	\vspace{-0.1cm}
	\setcounter{subfigure}{0}
	\subfloat[AuCs for PCR (top) and mIoU@$k$ (bottom) curves on the PF-PASCAL.]{
	\includegraphics[width=0.95\textwidth]{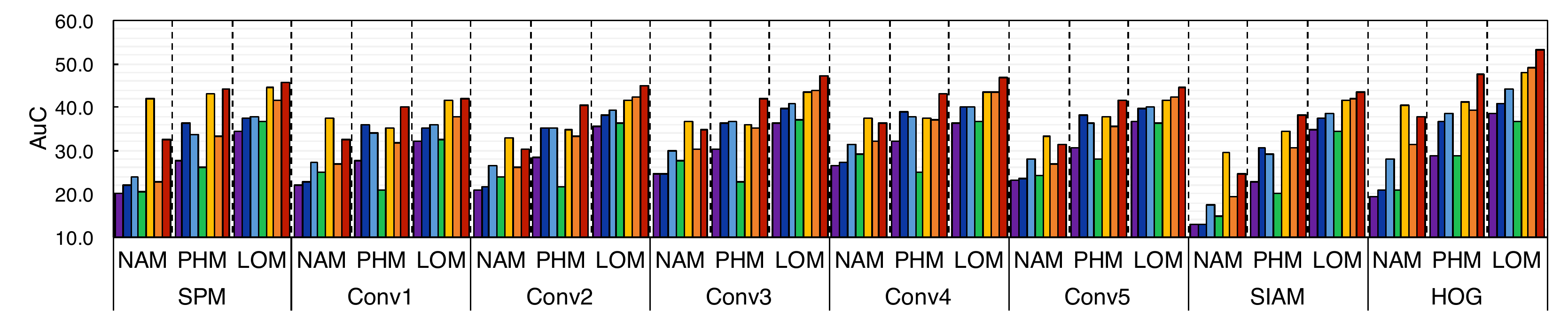}}

	\subfloat{
	\includegraphics[width=0.95\textwidth]{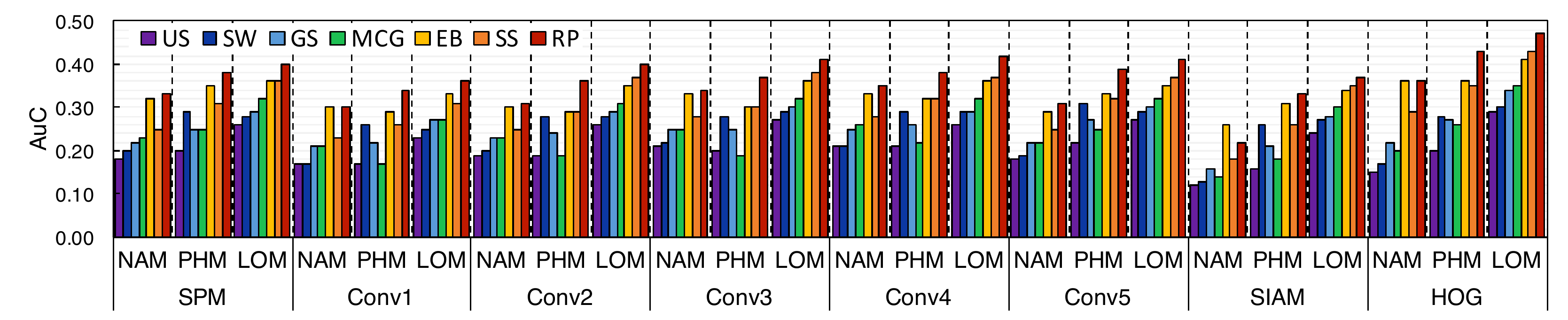}}
	\vspace{-0.1cm}
	\setcounter{subfigure}{1}
	\subfloat[AuCs for PCR (top) and mIoU@$k$ (bottom) curves on the PF-WILLOW.]{
	\includegraphics[width=0.95\textwidth]{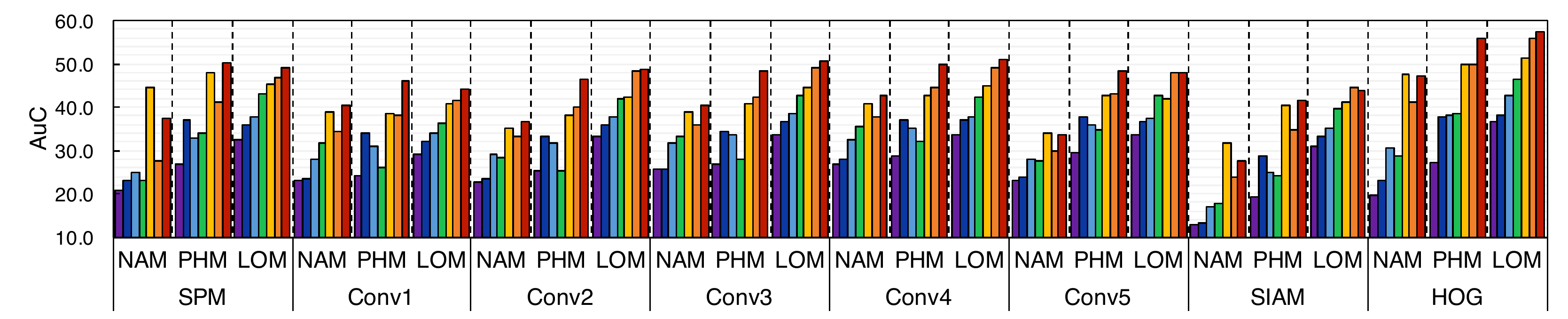}}
	\vfill

\caption{PF benchmark evaluation on AuCs for PCR and mIoU@$k$ plots: (a) PF-PASCAL dataset, and (b) PF-WILLOW dataset. We can see that combining LOM, RP, and HOG performs best in both metrics and datasets, and the challenging PF-PASCAL dataset shows slightly lower matching precision and retrieval accuracy than the PF-WILLOW. \textbf{(Best viewed in color.)}}
\label{fig:benchmark-auc}	
\end{figure*}

\begin{table*}[h!]
\centering
\footnotesize
\caption{AuC performance for PCR plots on the PF-PASCAL dataset (RP w/ HOG).}
\vspace{-0.3cm}
\addtolength{\tabcolsep}{-3.0pt}
\begin{tabular}{l c c c c c c c c c c c c c c c c c c c c c c}
\toprule
\multicolumn{1}{c}{Methods} & aero & bike & bird & boat & bot & bus & car & cat & cha & cow & tab & dog & hor & mbik & pers & plnt & she & sofa & trai& tv & Avg.\\
\midrule
\midrule
					 					    LOM	&{0.52}	&{0.56}	&{0.34}	&{0.39}	&{0.47}	&{0.61}	&{0.58}	&{0.34}	&{0.43}	&{0.43}	&{0.27}	&{0.36}	&{0.46}	&{0.48}	&{0.31}	&{0.34}	&{0.35}	&{0.37}	&{0.52}	&0.50	&{0.43}\\
\cmidrule{1-22}
Upper bound &0.70	&0.72	&0.63	&0.66	&0.71	&0.77	&0.73	&0.63	&0.72	&0.69	&0.57	&0.67	&0.70	&0.72	&0.66	&0.62	&0.53	&0.65	&0.73	&0.78	&0.68\\
\bottomrule
\end{tabular}	
\label{tb:pf_auc_pcr_iou_per_cls_PASCAL}
\end{table*}

\textbf{Qualitative comparison.} Figure \ref{fig:gt_generation}(e-g) shows a qualitative comparison between region matching algorithms on a pair of images and depicts correct matches found by each variant of proposal flow. In this example, at the IoU threshold $0.5$, the numbers of correct matches are 16, 5, and 38 for NAM, PHM~\cite{cho2015unsupervised}, and LOM, respectively. This shows that PHM may give worse performance than even NAM when there is much clutter in background. In contrast, the local regularization in LOM alleviates the effect of such clutter.

\textbf{Quantitative comparison on PF-PASCAL.} Figure \ref{fig:benchmark} summarizes the matching and retrieval performance on average for all object classes with a variety of combination of object proposals, feature descriptors, and matching algorithms. Figure \ref{fig:benchmark}(a) compares different types of object proposals with fixed matching algorithm and feature descriptor (LOM w/ HOG). RP gives the best matching precision and retrieval accuracy among the object proposals. 
An upper bound on precision is measured for object proposals (around a given object) in the image $\mathcal{I}$ using corresponding ground truths in image $\mathcal{I}^\prime$, that is the best matching accuracy we can achieve with each proposal method. To this end, for each region $r$ in the image $\mathcal{I}$, we find the region $r^\prime$ in the image $\mathcal{I}^\prime$ that has the highest IoU score given the region $r$'s ground-truth correspondence $r^\star$ in the image $\mathcal{I}^\prime$, and use the score as an upper bound precision. The upper bound (UB) plots show that RP generates more consistent regions than other proposal methods, and is adequate for region matching. RP shows higher matching precision than other proposals especially when the IoU threshold~$\tau$ is low. 
The evaluation results for different features (LOM w/ RP) are shown in Fig.~\ref{fig:benchmark}(b). The HOG descriptor gives the best performance in matching and retrieval. The CNN features in our comparison come from AlexNet~\cite{krizhevsky2012imagenet} trained for ImageNet classification.~Such CNN features have a task-specific bias to capture discriminative parts for classification, which may be less adequate for patch correspondence or retrieval than engineered features such as HOG. Similar conclusions are found in recent papers~\cite{long2014convnets, paulin2015localetal}. See, for example, Table 3 in~\cite{paulin2015localetal} where SIFT outperforms all AlexNet features (Conv1-5). Among ConvNet features, the fourth and first convolutional layers (Conv4 and Conv1) show the best and worst performance, respectively, while other layers perform similar to SPM. This confirms the finding in~\cite{zagoruyko2015learning}, which shows that Conv4 gives the best matching performance among ImageNet-trained ConvNet features. The SIAM feature is designed to compute patch similarity, and thus it can be used as a replacement for any task involving SIFT. This type of feature descriptor using Siamese or triplet networks such as~\cite{zagoruyko2015learning, han2015matchnet, simo2015discriminative} works well in finding correspondences between images containing the same object with moderate view point changes, e.g., as in the stereo matching task. But, we can see that this feature descriptor is less adequate for semantic flow, i.e., finding correspondences of different scenes and objects. The main reason is that the training dataset~\cite{brown2011discriminative} does not feature intra-class variations. We will show that the dense version of our proposal flow also outperforms a learning-based semantic flow method in Section~\ref{subsec:ex-flow}. Figure \ref{fig:benchmark}(c) compares the performance of different matching algorithms (RP w/ HOG), and shows that LOM outperforms others in matching as well as retrieval. 

Figure \ref{fig:benchmark-auc}(a) shows the area under curve (AuC) for PCR (top) and mIoU@$k$ (bottom) plots on average for all object classes with all combinations of object proposals, feature descriptors, and matching algorithms. This suggests that combining LOM, RP, and HOG performs best in both metrics. In Table~\ref{tb:pf_auc_pcr_iou_per_cls_PASCAL}, we show AuCs of PCR plots for each class of the PF-PASCAL dataset (RP w/ HOG). We can see that rigid objects~(e.g.,~bus and car) show higher matching precision than deformable ones~(e.g.,~person and bird).

\begin{figure*}[h!]
\centering
	\subfloat{
	\includegraphics[width=0.95\textwidth]{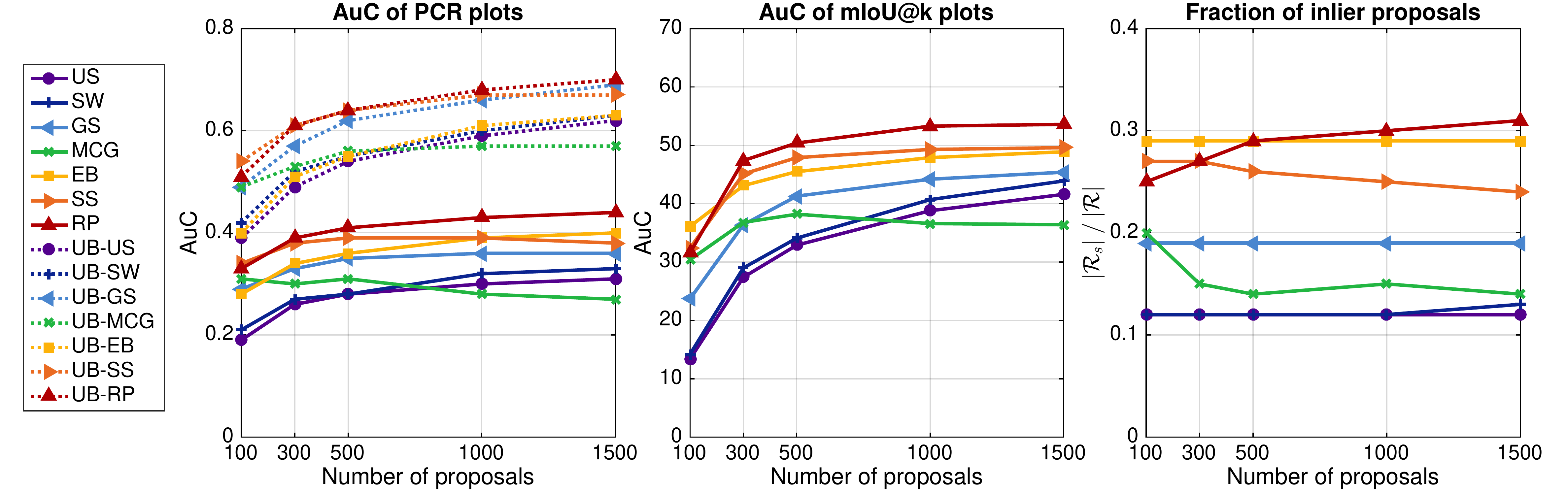}}

	\subfloat{
	\includegraphics[width=0.95\textwidth]{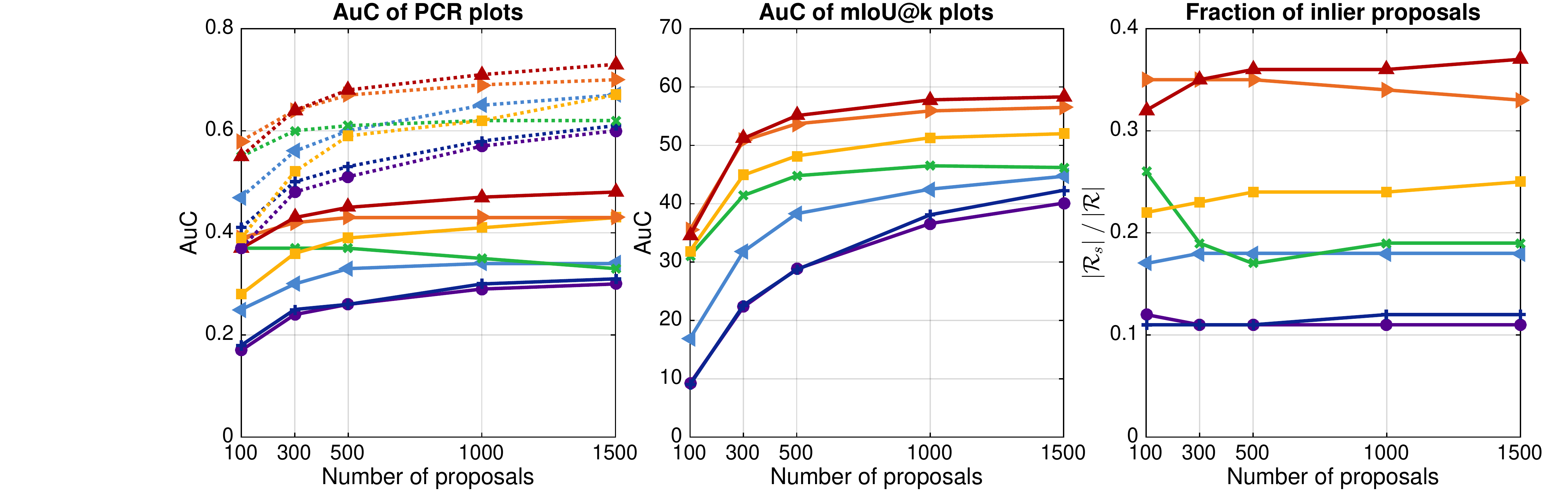}}
	
	\vfill
\caption{AuCs for PCR and mIoU$@$k plots and fraction of inlier proposals over all proposals on the PF-PASCAL (top) and PF-WILLOW (bottom). We can see that matching precision (left, PCR plots) and retrieval accuracy (center, mIoU$@$k plots) are slightly increasing, except MCG. The MCG is designed to obtain high precision with small number of proposals, so the fraction $|\mathcal{R}_s| / |\mathcal{R}|$ (right) decreases as the number of proposals. The LOM method is used for region matching with the HOG descriptor. \textbf{(Best viewed in color.)}}
\label{fig:benchmark-numProp_WILLOW}	
\end{figure*}

\begin{table*}[h!]
\centering
\footnotesize
\caption{AuC performance for PCR plots on the PF-WILLOW dataset (RP w/ HOG).}
\vspace{-0.3cm}
\addtolength{\tabcolsep}{-2.0pt}
\begin{tabular}{l c c c c c c c c c c c }
\toprule
\multicolumn{1}{c}{Methods} & car(S) & car(G) & car(M) & duc(S) & mot(S) & mot(G) & mot(M) & win(w/o C) & win(w/ C) & win(M) & Avg. \\
\midrule
\midrule
					 					    LOM	& {0.61} & 	{0.50}	&   {0.45}&	{0.50}	&  {0.42}& 	{0.40}&	{0.35}&	{0.69}&	0.30&	0.47&	{0.47}\\
\cmidrule{1-12}
Upper bound & 0.75&	0.69&	0.69&	0.72&	0.70&	0.70&	0.67&	0.80&	0.68&	0.73&	0.71\\
\bottomrule
\end{tabular}	
\label{tb:pf_auc_pcr_per_cls_WILLOW}
\end{table*}

\textbf{Quantitative comparison on PF-WILLOW.} We perform the same experiments with the PF-WILLOW dataset: The behavior of the average matching and retrieval performance is almost the same as the one for the PF-PASCAL dataset shown in Fig.~\ref{fig:benchmark}, so we omit these results. They can be found on our project webpage for completeness. In Fig~\ref{fig:benchmark-auc}(b), we show the AuC for PCR (top) and mIoU@$k$ (bottom) plots. We have the same conclusion here in the PF-PASCAL dataset (Fig~\ref{fig:benchmark-auc}(a)), but we can achieve higher matching precision and retrieval accuracy than for the challenging PF-PASCAL dataset.  In Table \ref{tb:pf_auc_pcr_per_cls_WILLOW}, we show AuCs of PCR plots for each sub-class. From this table, we can see that 1) higher matching precision is achieved with objects having a similar pose (e.g., mot(S)~vs.~mot(M)), 2) performance decreases for deformable object matching (e.g., duck(S)~vs.~car(S)), and 3) matching precision can increase drastically by eliminating background clutters (e.g., win(w/o C)~vs.~win(w/ C)), which verifies out motivation of using object proposals for semantic flow.

\textbf{Effect of the number of proposals.}
In Fig.~\ref{fig:benchmark-numProp_WILLOW}, we show the AuCs of PCR (left) and mIoU@$k$ (center) plots, on the PF-PASCAL (top) and PF-WILLOW (bottom), as a function of the number of object proposals. We see that 1) upper bounds on matching precision of all proposals are continuously growing, except MCG, as the number of proposal increases, and 2) matching precision and retrieval accuracy of proposal flow are increasing as well, but at a slightly slower rate. On the one hand, as the number of proposals is increasing, the number of inlier proposals, i.e., regions around object bounding boxes $|\mathcal{R}_s|$, is increasing, and thus we can achieve a higher upper bound. On the other hand, the number of outlier proposals, i.e., $| \mathcal{R}|$ - $|\mathcal{R}_s|$, is increasing as well, which prevents us from finding correct matches. Overall, matching precision and retrieval accuracy increase with the number of proposals (except for MCG), and start to saturate around 1000 proposals. We hypothesize that this is related to the fraction of inliers over all proposals, i.e.,  $|\mathcal{R}_s| / |\mathcal{R}|$, which may decrease in the case of MCG. To verify this, we plot this fraction as a function of the number of object proposals~(Fig.~\ref{fig:benchmark-numProp_WILLOW}, right). We can see that the fraction of MCG is drastically decreasing as the number of proposals, which means that MCG generates more and more outlier proposals corresponding, e.g., to background clutter. The reason is that high recall is the main criteria when designing most object proposal methods, but MCG is designed to achieve high precision with a small number of proposals~\cite{arbelaez2014multiscale}. 

\subsection{Flow field}\label{subsec:ex-flow}
To compare our method with state-of-the-art semantic flow methods, we compute a dense flow field from our proposal flows (Section~\ref{subsec:flowfield}), and evaluate image alignment between all pairs of images in each subset of the PF-PASCAL and PF-WILLOW datasets. We also compare the matching accuracy with existing datasets: Clatech-101~\cite{fei2006one}, PASCAL parts~\cite{zhou2015flowweb}, and Taniai's datasets~\cite{taniai2016joint}. In each case, we compare the proposal flow to the state of the art. For proposal flow, we use a SS method and HOG descriptors, unless otherwise specified, and use publicly available codes for all compared methods. 

\begin{figure*}
\centering
	
	\subfloat{
	\includegraphics[width=0.135\textwidth, valign=c,frame]{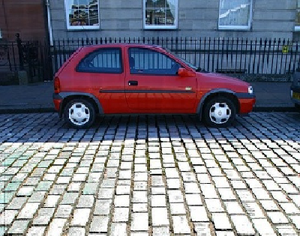}
	\includegraphics[width=0.135\textwidth, valign=c,frame]{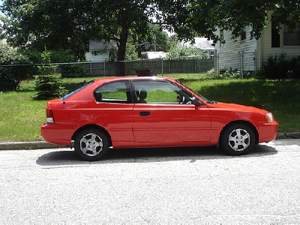}
	\includegraphics[width=0.135\textwidth, valign=c,frame]{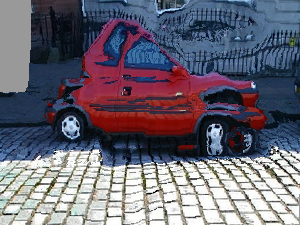}
	\includegraphics[width=0.135\textwidth, valign=c,frame]{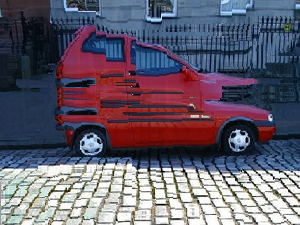}
	\includegraphics[width=0.135\textwidth, valign=c,frame]{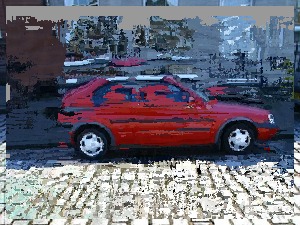}
	\includegraphics[width=0.135\textwidth, valign=c,frame]{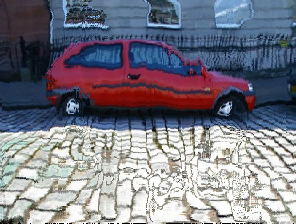}
	\includegraphics[width=0.135\textwidth, valign=c,frame]{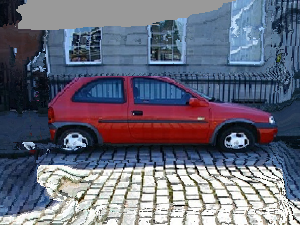}}
	\vspace{-0.3cm}

	\subfloat{
	\includegraphics[width=0.135\textwidth, valign=c,frame]{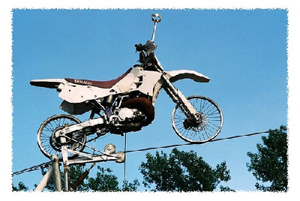}
	\includegraphics[width=0.135\textwidth, valign=c,frame]{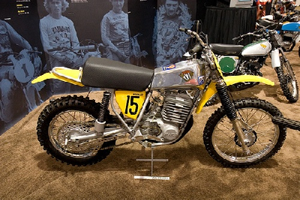}
	\includegraphics[width=0.135\textwidth, valign=c,frame]{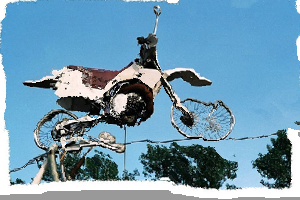}
	\includegraphics[width=0.135\textwidth, valign=c,frame]{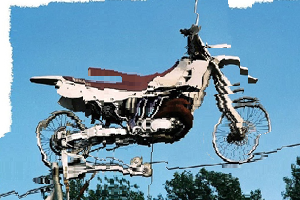}
	\includegraphics[width=0.135\textwidth, valign=c,frame]{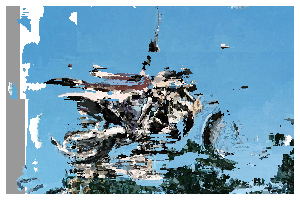}
	\includegraphics[width=0.135\textwidth, valign=c,frame]{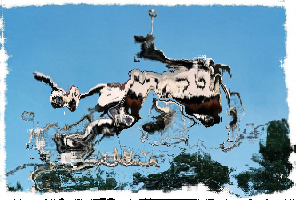}
	\includegraphics[width=0.135\textwidth, valign=c,frame]{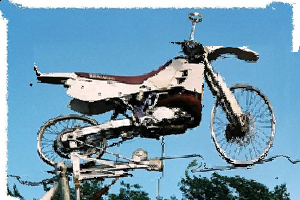}}	
	\vspace{-0.3cm}

	\subfloat{
	\includegraphics[width=0.135\textwidth, valign=c,frame]{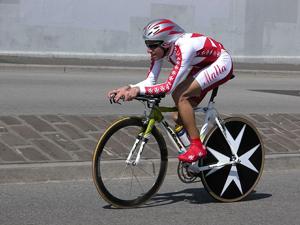}
	\includegraphics[width=0.135\textwidth, valign=c,frame]{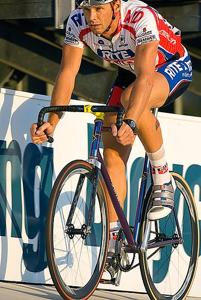}
	\includegraphics[width=0.135\textwidth, valign=c,frame]{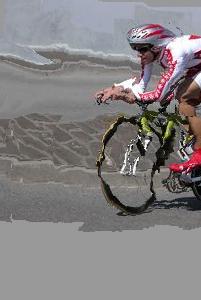}
	\includegraphics[width=0.135\textwidth, valign=c,frame]{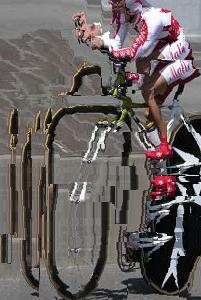}
	\includegraphics[width=0.135\textwidth, valign=c,frame]{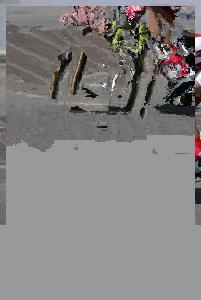}
		\includegraphics[width=0.135\textwidth, valign=c,frame]{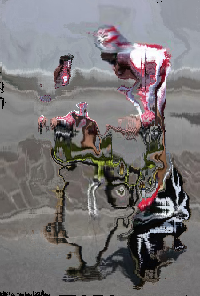}
	\includegraphics[width=0.135\textwidth, valign=c,frame]{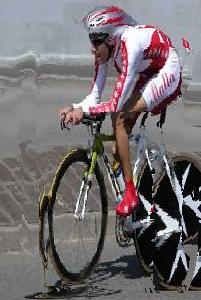}}
	\vspace{-0.3cm}

	\subfloat{
	\includegraphics[width=0.135\textwidth, valign=c,frame]{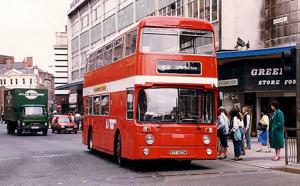}
	\includegraphics[width=0.135\textwidth, valign=c,frame]{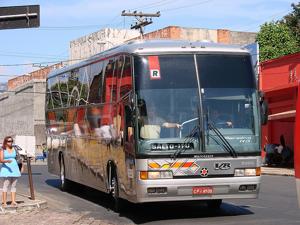}
	\includegraphics[width=0.135\textwidth, valign=c,frame]{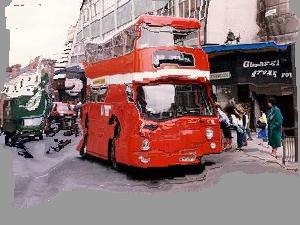}
	\includegraphics[width=0.135\textwidth, valign=c,frame]{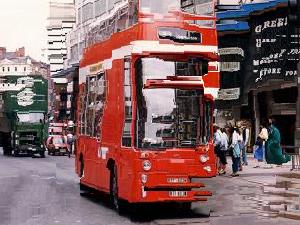}
	\includegraphics[width=0.135\textwidth, valign=c,frame]{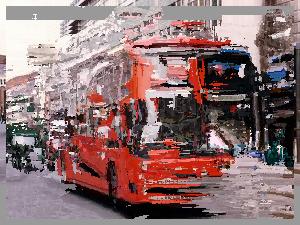}
	\includegraphics[width=0.135\textwidth, valign=c,frame]{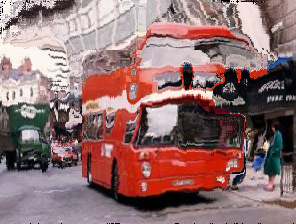}
	\includegraphics[width=0.135\textwidth, valign=c,frame]{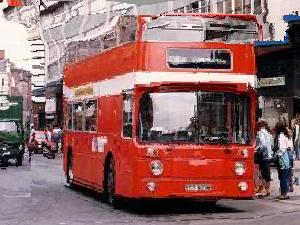}}
	\vspace{-0.3cm}

	\subfloat{
	\includegraphics[width=0.135\textwidth, valign=c,frame]{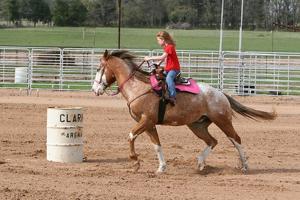}
	\includegraphics[width=0.135\textwidth, valign=c,frame]{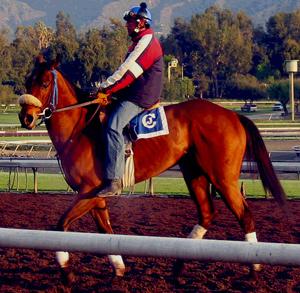}
	\includegraphics[width=0.135\textwidth, valign=c,frame]{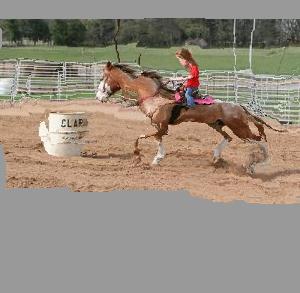}
	\includegraphics[width=0.135\textwidth, valign=c,frame]{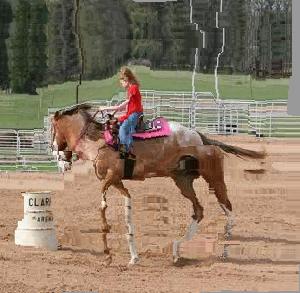}
	\includegraphics[width=0.135\textwidth, valign=c,frame]{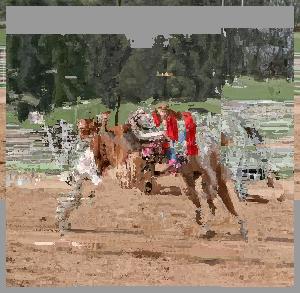}
	\includegraphics[width=0.135\textwidth, valign=c,frame]{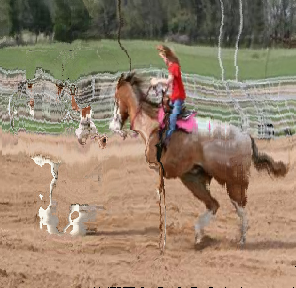}
	\includegraphics[width=0.135\textwidth, valign=c,frame]{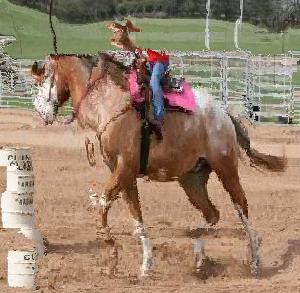}}
	\vspace{-0.3cm}

	\setcounter{subfigure}{0}
	\subfloat[Source image.]{
	\includegraphics[width=0.135\textwidth, valign=c,frame]{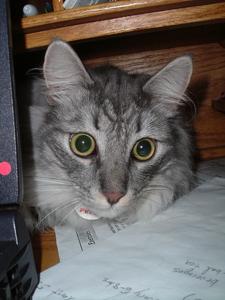}}
	\subfloat[Target image.]{
	\includegraphics[width=0.135\textwidth, valign=c,frame]{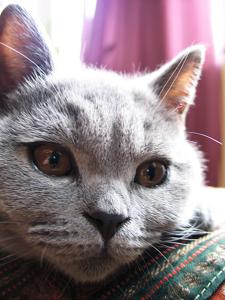}}
	\subfloat[DeepFlow.]{
	\includegraphics[width=0.135\textwidth, valign=c,frame]{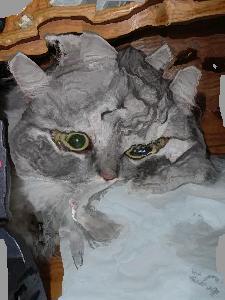}}
	\subfloat[SIFT Flow.]{
	\includegraphics[width=0.135\textwidth, valign=c,frame]{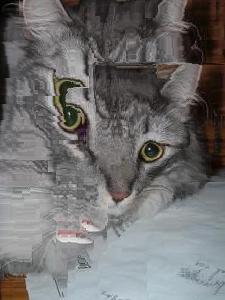}}
	\subfloat[DSP.]{
	\includegraphics[width=0.135\textwidth, valign=c,frame]{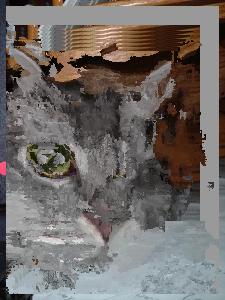}}
	\subfloat[Zhou \emph{et al}.]{
	\includegraphics[width=0.135\textwidth, valign=c,frame]{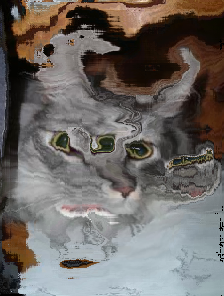}}
	\subfloat[Proposal Flow.]{
	\includegraphics[width=0.135\textwidth, valign=c,frame]{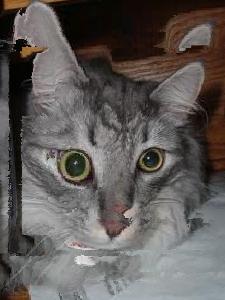}}
\vfill
\caption{Examples of dense flow field.~(a-b) Sourse images are warped to the target images using the dense correspondences estimated by (c) DeepFlow~\cite{weinzaepfel2015deepmatching}, (d) SIFT Flow~\cite{liu2011sift}, (e) DSP~\cite{kim2013deformable}, (f) Zhou \emph{et al}.~\cite{zhou2016learning}, and (g) Proposal Flow (LOM w/ RP and HOG). Compared to the existing methods, proposal flow is robust to background clutter, and translation and scale changes between objects. The first two images are from the PF-WILLOW and remaining ones are from the PF-PASCAL.}
\label{fig:correspondence}	
\end{figure*}

\begin{figure*}[h!]
\centering
	\subfloat{
	\includegraphics[width=0.135\textwidth, frame]{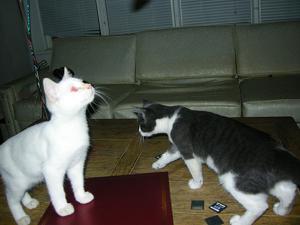}
	\includegraphics[width=0.135\textwidth, frame]{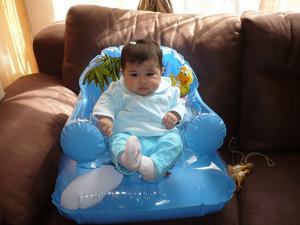}
	\includegraphics[width=0.135\textwidth, frame]{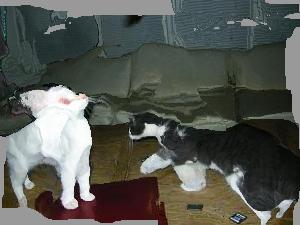}
	\includegraphics[width=0.135\textwidth, frame]{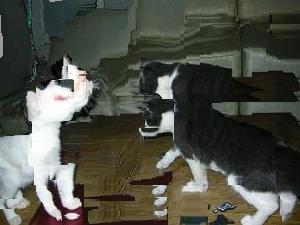}
	\includegraphics[width=0.135\textwidth, frame]{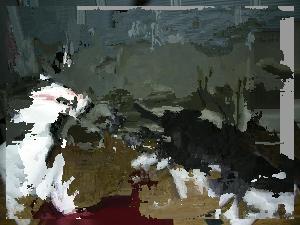}
	\includegraphics[width=0.135\textwidth, frame]{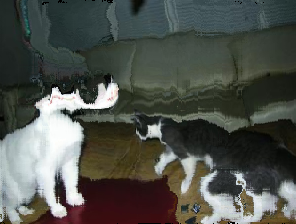}
	\includegraphics[width=0.135\textwidth, frame]{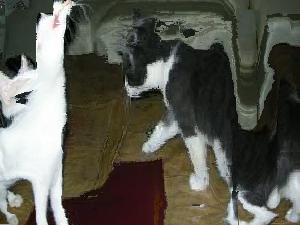}}
	\vspace{-0.3cm}

	\subfloat{
	\includegraphics[width=0.135\textwidth, valign=c, frame]{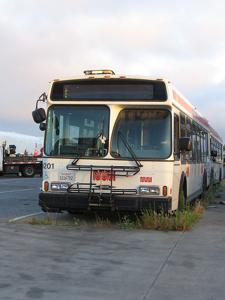}
	\includegraphics[width=0.135\textwidth, valign=c, frame]{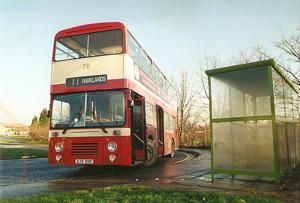}
	\includegraphics[width=0.135\textwidth, valign=c, frame]{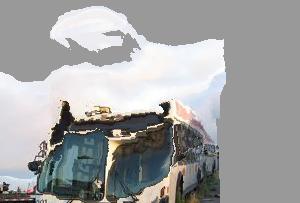}
	\includegraphics[width=0.135\textwidth, valign=c, frame]{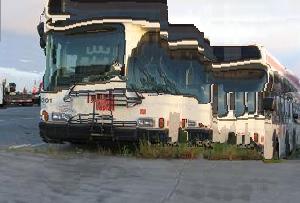}
	\includegraphics[width=0.135\textwidth, valign=c, frame]{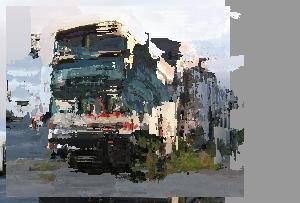}
	\includegraphics[width=0.135\textwidth, valign=c, frame]{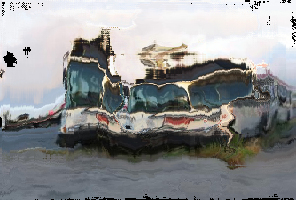}
	\includegraphics[width=0.135\textwidth, valign=c, frame]{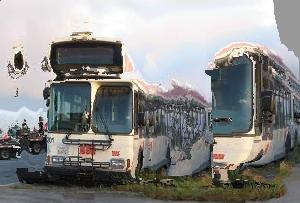}}
	\vspace{-0.3cm}

	\setcounter{subfigure}{0}
	\subfloat[Source image.]{
	\includegraphics[width=0.135\textwidth, valign=c, frame]{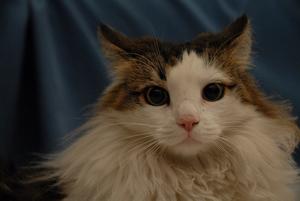}}
	\subfloat[Target image.]{
	\includegraphics[width=0.135\textwidth, valign=c,frame]{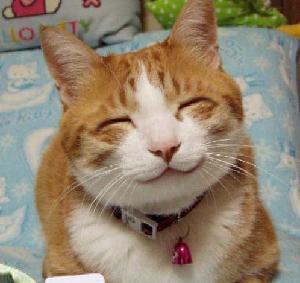}}
	\subfloat[DeepFlow.]{
	\includegraphics[width=0.135\textwidth, valign=c,frame]{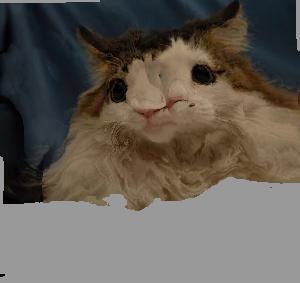}}
	\subfloat[SIFT Flow.]{
	\includegraphics[width=0.135\textwidth, valign=c,frame]{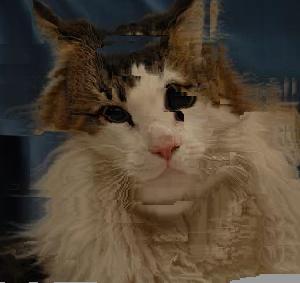}}
	\subfloat[DSP.]{
	\includegraphics[width=0.135\textwidth, valign=c,frame]{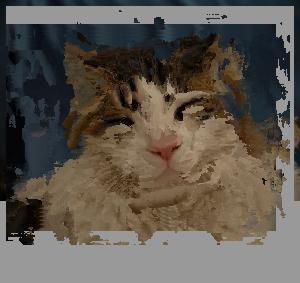}}
	\subfloat[Zhou \emph{et al}.]{
	\includegraphics[width=0.135\textwidth, valign=c,frame]{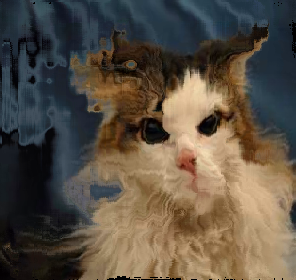}}
	\subfloat[Proposal Flow.]{
	\includegraphics[width=0.135\textwidth, valign=c,frame]{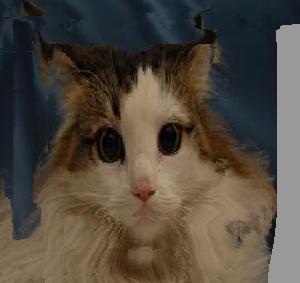}}
	\vfill	

\caption{Failure examples of (from top to bottom) sofa, bus, and cat classes on the PF-PASCAL dataset. (a-b) Source images are warped to the target images using the dense correspondences estimated by (c) DeepFlow~\cite{weinzaepfel2015deepmatching}, (d) SIFT Flow~\cite{liu2011sift}, (e) DSP~\cite{kim2013deformable}, (f) Zhou \emph{et al}.~\cite{zhou2016learning} and (g) Proposal Flow (LOM w/ RP and HOG). Proposal flow is hard to deal with images containing (from top to bottom) severe occlusion, similarly shaped objects, and deformation.}
\label{fig:vis_corres_failure_PASCAL}	
\end{figure*}

\begin{table}[ht!]
\centering
\caption{PCK~($\alpha = 0.1$) comparison for dense flow field on the PF dataset~(PF-PASCAL / PF-WILLOW).}
\vspace{-0.3cm}
\addtolength{\tabcolsep}{-3.0pt}
\begin{tabular}{l c c c c c}
\toprule
\multicolumn{1}{c}{Methods} & SW~\cite{hosang2015what} & MCG~\cite{arbelaez2014multiscale} & EB~\cite{zitnick2014edge} & SS~\cite{uijlings2013selective} & RP~\cite{manen2013prime}\\
\midrule
\midrule
NAM &  0.29/0.44  &  0.27/0.46 & \textbf{0.37}/\textbf{0.51}&  0.36/0.52 & 0.37/0.54\\
PHM  & \textbf{0.37}/\textbf{0.48}      &0.35/0.48&	0.35/0.45& 0.42/0.55& 0.42/0.54\\
LOM &  0.35/0.42	     &\textbf{0.38}/ \textbf{0.49}& \textbf{0.37}/0.45&\textbf{0.45}/\textbf{0.56}&\textbf{0.44}/\textbf{0.55}\\
\midrule
DeepFlow~\cite{weinzaepfel2015deepmatching} & \multicolumn{4}{c}{0.21/0.20}\\
GMK~\cite{duchenne2011graph} & \multicolumn{4}{c}{0.27/0.27}\\
SIFT Flow~\cite{liu2011sift} & \multicolumn{4}{c}{0.33/0.38}\\
DSP~\cite{kim2013deformable} & \multicolumn{4}{c}{0.30/0.37}\\
Zhou~\emph{et al}.~\cite{zhou2016learning} & \multicolumn{4}{c}{0.30/0.41}\\
\bottomrule
\end{tabular}
\label{tb:pck}
\end{table}

\begin{table*}[h!]
\centering
\footnotesize
\caption{PCK ($\alpha=0.1$) comparison for dense flow field on the PF-PASCAL dataset~(SS w/ HOG).}
\vspace{-0.3cm}
\addtolength{\tabcolsep}{-3.0pt}
\begin{tabular}{l c c c c c c c c c c c c c c c c c c c c c}
\toprule
\multicolumn{1}{c}{Methods} & aero & bike & bird & boat & bot & bus & car & cat & cha & cow & tab & dog & hor & mbik & pers & plnt & she & sofa & trai& tv & Avg.\\
\midrule
\midrule
LOM& \textbf{0.75}	&\textbf{0.76}	&\textbf{0.34}	&\textbf{0.41}	&\textbf{0.55}	&\textbf{0.71}	&\textbf{0.73}	&\textbf{0.32}	&\textbf{0.41}	&\textbf{0.41}	&0.21	&\textbf{0.27}	&\textbf{0.38}	&\textbf{0.57}	&\textbf{0.29}	&\textbf{0.17}	&0.33	&\textbf{0.34}	&\textbf{0.54}	&\textbf{0.46}	&\textbf{0.45}\\
\midrule
\multicolumn{1}{l}{DeepFlow~\cite{weinzaepfel2015deepmatching}} &0.55	&0.31	&0.10	&0.19	&0.24	&0.36	&0.31	&0.12	&0.22	&0.10	&0.23	&0.07	&0.11	&0.32	&0.10	&0.08	&0.07	&0.20	&0.31	&0.17	&0.21\\
\multicolumn{1}{l}{GMK~\cite{duchenne2011graph}} &0.61	&0.49	&0.15	&0.21	&0.29	&0.47	&0.52	&0.14	&0.23	&0.23	&0.24	&0.09	&0.13	&0.39	&0.12	&0.16	&0.10	&0.22	&0.33	&0.22	&0.27\\
\multicolumn{1}{l}{SIFT Flow	\cite{liu2011sift}}  &0.61	&0.56	&0.20	&0.34	&0.32	&0.54	&0.56	&0.26	&0.29	&0.21	&\textbf{0.33}	&0.17	&0.23	&0.43	&0.18	&0.17	&0.17	&0.31	&0.41	&0.34	&0.33\\   
\multicolumn{1}{l}{DSP~\cite{kim2013deformable}} &0.64	&0.56	&0.17	&0.27	&0.38	&0.51	&0.55	&0.20	&0.23	&0.24	&0.19	&0.15	&0.23	&0.41	&0.15	&0.11	&0.18	&0.27	&0.35	&0.28	&0.30\\  
\multicolumn{1}{l}{Zhou~\emph{et al}.~\cite{zhou2016learning}} & 0.58	& 0.35 &	0.15 &	0.27 &	0.36	& 0.40 &	0.42 &	0.23 &	0.26 &	0.29 &	0.22 & 	0.20 & 	0.13 &	0.33 &	0.16 &	0.18	&  \textbf{0.48} & 	0.27 &	0.34 &	0.28 &	0.30\\
  
\bottomrule
\end{tabular}	
\label{tb:pf_pck_per_cls_PASCAL}
\end{table*}

\begin{table*}[h!]
\centering
\footnotesize
\caption{PCK ($\alpha=0.1$) comparison for dense flow field on the PF-WILLOW dataset~(SS w/ HOG).}
\vspace{-0.3cm}
\addtolength{\tabcolsep}{-2.0pt}
\begin{tabular}{l c c c c c c c c c c c}
\toprule
\multicolumn{1}{c}{Methods} & car(S) & car(G) & car(M) & duc(S) & mot(S) & mot(G) & mot(M) & win(w/o C) & win(w/ C) & win(M) & Avg. \\
\midrule
\midrule
LOM&\textbf{0.86}	&\textbf{0.60}	&\textbf{0.53}	&\textbf{0.64}	&\textbf{0.49}	&\textbf{0.25}	&\textbf{0.29}	&\textbf{0.91}	&\textbf{0.37}	&\textbf{0.65}	&\textbf{0.56}\\
\midrule
\multicolumn{1}{l}{DeepFlow~\cite{weinzaepfel2015deepmatching}} &0.33	&0.13	&0.22 &0.20	&0.20	&0.08	&0.13	&0.46	&0.08	&0.18	&0.20\\
\multicolumn{1}{l}{GMK~\cite{duchenne2011graph}} & 0.48	&0.25	&0.34	&0.27	&0.31	&0.12	&0.15	&0.41	&0.17	&0.18	&0.27\\
\multicolumn{1}{l}{SIFT Flow	\cite{liu2011sift}}           &0.54	&0.37	&0.36	&0.32	&0.41	&0.20   &0.23	&0.83	&0.16   &0.33   &0.38\\
\multicolumn{1}{l}{DSP~\cite{kim2013deformable}}     &0.46  &0.30   &0.32   &0.25   &0.31   &0.15   &0.14   &0.85   &0.25   &0.64   &0.37\\
\multicolumn{1}{l}{Zhou~\emph{et al}.~\cite{zhou2016learning}} &0.77 &0.34 &0.52 &0.42 &0.34 &0.19 &0.20 &0.78 &0.19 &0.38 &0.41\\
\bottomrule
\end{tabular}	
\label{tb:pf_pck_per_cls_WILLOW}
\end{table*}

\begin{table}[h!]
\centering
\footnotesize
\captionsetup{font={small}}
\caption{Runtime comparison for dense flow field on the PF-PASCAL dataset~(SS w/ HOG).}
\vspace{-0.3cm}
\begin{threeparttable}
\begin{tabular}{l c}
\toprule
\multicolumn{1}{c}{Methods} & Time (\emph{s})\\
\midrule
\midrule
NAM & 4.6 $\pm$ 1.0\\
PHM & 5.4 $\pm$ 1.1\\
LOM & 8.8 $\pm$ 1.3\\
\midrule
DeepFlow~\cite{weinzaepfel2015deepmatching} & 4.7 $\pm$ 0.6\tnote{$\dagger$}\\
GMK~\cite{duchenne2011graph} & 2.4 $\pm$ 0.3\tnote{$\dagger$}\\
SIFT Flow~\cite{liu2011sift} & 4.2 $\pm$ 0.8\tnote{$\dagger$}\\
DSP~\cite{kim2013deformable} & 4.8 $\pm$ 0.8\tnote{$\dagger$}\\
\bottomrule
\end{tabular}
\begin{tablenotes}
        \item[$\dagger$] \scriptsize{We used author provided \texttt{MEX} implementations.}
\end{tablenotes}
\end{threeparttable}
\label{tb:pf_runtime_PASCAL}
\end{table}

\textbf{Matching results on PF datasets.} We test five object proposal methods (SW, MCG, EB, SS, RP). For an evaluation metric, we use PCK between warped keypoints and ground-truth ones~\cite{long2014convnets, yang2013articulated}. Ground-truth keypoints are deemed to be correctly predicted if they lie within $\alpha \text{max}(h,w)$ pixels of the predicted points for $\alpha$ in $[0, 1]$, where $h$ and $w$ are the height and width of the object bounding box, respectively. Table \ref{tb:pck} shows the average PCK $(\alpha = 0.1)$ over all object classes. In our benchmark, all versions of proposal flow significantly outperform SIFT Flow~\cite{liu2011sift}, DSP~\cite{kim2013deformable}, and DeepFlow~\cite{weinzaepfel2015deepmatching}, and proposal flow with PHM and LOM gives better performance than the learning-based method~\cite{zhou2016learning}. LOM with SS or RP outperforms other combination of matching and proposal methods, which coincides with the results in Section \ref{subsec:ex-region-matching}. Tables \ref{tb:pf_pck_per_cls_PASCAL} and \ref{tb:pf_pck_per_cls_WILLOW} show the average PCK $(\alpha = 0.1)$ over each object class on the PF-PASCAL and PF-WILLOW, respectively. This shows that proposal flow consistently outperforms other methods for all object classes except for table and sheep classes in both datasets. We can also see that the learning-based method~\cite{zhou2016learning} does not generalize other object classes that are not contained in the PASCAL training set~(e.g., duc(S)), and are not robust to the outliers~(e.g., wine~(w/ c)). Figure~\ref{fig:correspondence} gives a qualitative comparison with the state of the art on the PF-WILLOW and PF-PASCAL datasets. The better alignment found by proposal flow here is clearly visible. Specifically, proposal flow is robust to clutter and translation and scale changes between objects. Figure~\ref{fig:vis_corres_failure_PASCAL} shows failure examples of (from top to bottom) sofa, bus, and cat classes on the PF-PASCAL dataset, where we see proposal flow does not handle image pairs that contain severe occlusion, objects having similar shape, and deformation. Our current (un-optimized) \texttt{MATLAB} implementation takes on average 8.8 seconds on 2.5 GHz CPU for computing dense flow field using LOM w/ SS and HOG. Table~\ref{tb:pf_runtime_PASCAL} shows runtime comparisons. 

\begin{table}
\centering
\caption{Matching accuracy on the Caltech-101 dataset~(HOG).}
\vspace{-0.3cm}
\addtolength{\tabcolsep}{-2.0pt}
\begin{tabular}{ l l c c c }
\toprule
\multicolumn{1}{c}{Proposals} & \multicolumn{1}{c}{Methods} & LT-ACC & IoU & LOC-ERR \\
\midrule
\midrule
\multirow{1}{*}{SW~\cite{hosang2015what}} & LOM & \textbf{0.78} & 0.47 & \textbf{0.25} \\
\midrule
\multirow{3}{*}{SS~\cite{uijlings2013selective}} & NAM & 0.68 & 0.44 & 0.41 \\
                    & PHM & 0.74 & 0.48 & 0.32 \\
					& LOM & \textbf{0.78} & \textbf{0.50} & \textbf{0.25} \\
\midrule
\multirow{3}{*}{RP~\cite{manen2013prime}} & NAM & 0.70 & 0.44 & 0.39\\
					& PHM & 0.75 & 0.48 & 0.31\\
					& LOM & \textbf{0.78} & \textbf{0.50} & 0.26\\
\midrule
\multicolumn{2}{l}{DeepFlow~\cite{weinzaepfel2015deepmatching}}  &0.74& 0.40& 0.34\\
\multicolumn{2}{l}{GMK~\cite{duchenne2011graph}}  & 0.77& 0.42&0.34\\
\multicolumn{2}{l}{SIFT Flow~\cite{liu2011sift}}        & 0.75 & 0.48 & 0.32\\
\multicolumn{2}{l}{DSP~\cite{kim2013deformable}}              & 0.77 & 0.47 & 0.35\\
\bottomrule
\end{tabular}
\label{tb:caltech}
\end{table}

\begin{table}
\centering
\caption{Matching accuracy on the Taniai's dataset~(SS w/ HOG).}
\vspace{-0.3cm}
\addtolength{\tabcolsep}{-2.0pt}
\begin{tabular}{ l c c c c }
\toprule
\multicolumn{1}{c}{Methods} & FG3DCar & JODS & PASCAL & Avg. \\
\midrule
\midrule
LOM & 0.79 & \textbf{0.65} & \textbf{0.53} & \textbf{0.66} \\
\midrule
DFF~\cite{yang2014daisy} & 0.50 & 0.30 & 0.22 & 0.31\\
DSP~\cite{kim2013deformable}  & 0.49 & 0.47 & 0.38 & 0.45\\
SIFT Flow~\cite{liu2011sift}   &   0.63 & 0.51 & 0.36 & 0.50\\
Zhou~\emph{et al}.~\cite{zhou2016learning}        &     0.72 & 0.51 & 0.44 & 0.56 \\
Taniai~\emph{et al}.~\cite{taniai2016joint} & \textbf{0.83} & 0.60 & 0.48 & 0.64 \\
\bottomrule
\end{tabular}
\label{tb:taniai}
\end{table}

\begin{table}[ht!]
\centering
\caption{Matching accuracy on the PASCAL parts (SS w/ HOG).}
\vspace{-0.3cm}
\addtolength{\tabcolsep}{-2.0pt}
\begin{tabular}{l c c }
\toprule
\multicolumn{1}{c}{Methods} & IoU & PCK\\
\midrule
\midrule
NAM								& 0.35	&0.13\\
PHM								& 0.39	&\textbf{0.17}\\
LOM 								& \textbf{0.41} &\textbf{0.17}\\
\midrule
Congealing~\cite{learned2006data} 				& 0.38 	&0.11\\		    
RASL~\cite{peng2012rasl}          				& 0.39	&0.16\\
CollectionFlow~\cite{kemelmacher2012collection} & 0.38	&0.12\\
DSP~\cite{kim2013deformable}						& 0.39	&\textbf{0.17}\\
\midrule
FlowWeb~\cite{zhou2015flowweb}						& \textbf{0.43}	&\textbf{0.26}\\
\bottomrule
\end{tabular}
\label{tb:voc}
\end{table}

\textbf{{Matching results on Caltech-101.}}
We evaluate our approach on the Caltech-101
dataset~\cite{fei2006one}. Following the experimental protocol
in~\cite{kim2013deformable}, we randomly select 15 pairs of images for
each object class, and evaluate matching accuracy with three metrics:~Label transfer accuracy (LT-ACC)~\cite{liu2011nonparametric}, the IoU
metric, and the localization error (LOC-ERR) of corresponding pixel
positions. For LT-ACC, we transfer the class label of one image to the
other using dense correspondences, and count the number of correctly
labeled pixels. Similarly, the IoU score is measured between the
transferred label and ground truth.  Table \ref{tb:caltech} compares
quantitatively the matching accuracy of proposal flow to the state of
the art. It shows that proposal flow using LOM outperforms other
approaches, especially for the IoU score and the LOC-ERR of dense
correspondences. Note that compared to LT-ACC, these metrics evaluate
the matching quality for the foreground object, separate from 
irrelevant scene clutter. Our results verify that proposal flow
focuses on regions containing objects rather than scene clutter
and distracting details, enabling robust image matching against
outliers. %

\textbf{Matching results on Taniai's Benchmark.} We also evaluate flow accuracy on the dataset provided by~\cite{taniai2016joint} that consists of 400 image pairs of three groups: FG3DCar~(195 image pairs of vehicles from~\cite{lin2014jointly}), JODS~(81 image pairs of airplanes, horses, and cars from~\cite{rubinstein2013unsupervised}), and PASCAL~(124 image pairs of bicycles, motorbikes, buses, cars, trains from~\cite{hariharan2011semantic}). Matching accuracy is measured by the percentage of pixels in the ground-truth foreground region that have an error measure below a certain threshold. To this end, we compute the Euclidean distance between estimated and true flow vectors in a normalized scale where the larger dimensions of images are 100 pixels. Here, we use a threshold of 5 pixels following the work of~\cite{taniai2016joint}. We summarize average matching accuracy for each group in the Table~\ref{tb:taniai}. The method of~\cite{zhou2016learning} uses convolutional neural networks~(CNNs) to learn dense correspondence. Since there is no previous dataset available for training the networks for semantic flow, it leverages a 3D model to use the known synthetic-to-synthetic matches as ground truth, allowing cycle consistency to propagate the correct match information from synthetic to real images. The method of~\cite{taniai2016joint} leverages an additional cosegmentation to estimate dense correspondence. This is a similar idea to ours in that excluding background regions when estimating correspondences improves the matching accuracy. In the FG3DCar dataset, this method~\cite{taniai2016joint} shows better performance then ours. But, overall, our method achieves the best performance on average over all datasets, and even outperforms the learning based method of~\cite{zhou2016learning}.

\begin{figure*}[ht!]
\centering
	\subfloat{
	\includegraphics[width=0.12\textwidth, frame]{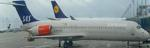}
	\includegraphics[width=0.12\textwidth, frame]{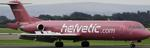}
	\includegraphics[width=0.12\textwidth, frame]{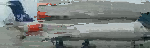}
	\includegraphics[width=0.12\textwidth, frame]{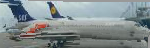}
	\includegraphics[width=0.12\textwidth, frame]{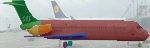}
	\includegraphics[width=0.12\textwidth, frame]{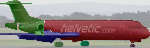}
	\includegraphics[width=0.12\textwidth, frame]{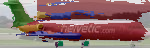}
	\includegraphics[width=0.12\textwidth, frame]{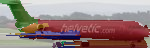}}	
	\vspace{-0.3cm}

	\subfloat{
	\includegraphics[width=0.12\textwidth, frame]{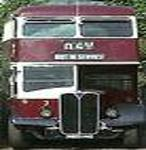}
	\includegraphics[width=0.12\textwidth, frame]{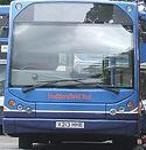}
	\includegraphics[width=0.12\textwidth, frame]{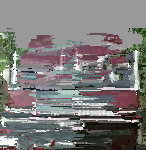}
	\includegraphics[width=0.12\textwidth, frame]{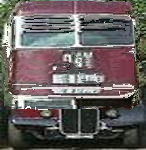}
	\includegraphics[width=0.12\textwidth, frame]{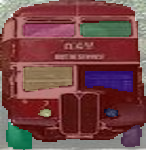}
	\includegraphics[width=0.12\textwidth, frame]{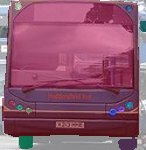}
	\includegraphics[width=0.12\textwidth, frame]{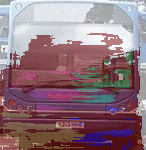}
	\includegraphics[width=0.12\textwidth, frame]{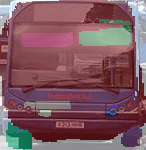}}	
	\vspace{-0.3cm}

	\subfloat{
	\includegraphics[width=0.12\textwidth, frame]{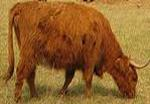}
	\includegraphics[width=0.12\textwidth, frame]{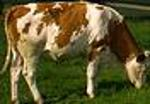}
	\includegraphics[width=0.12\textwidth, frame]{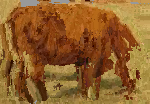}
	\includegraphics[width=0.12\textwidth, frame]{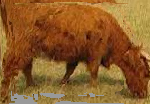}
	\includegraphics[width=0.12\textwidth, frame]{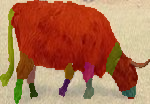}
	\includegraphics[width=0.12\textwidth, frame]{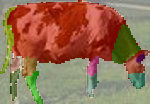}
	\includegraphics[width=0.12\textwidth, frame]{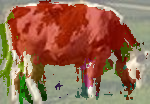}
	\includegraphics[width=0.12\textwidth, frame]{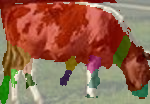}}
	\vspace{-0.3cm}

	\setcounter{subfigure}{0}
	\subfloat[\footnotesize{Source image.}]{
	\includegraphics[width=0.12\textwidth]{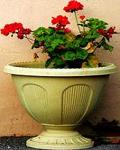}}
	\subfloat[\footnotesize{Target image.}]{
	\includegraphics[width=0.12\textwidth]{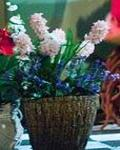}}
	\subfloat[\footnotesize{DSP.}]{
	\includegraphics[width=0.12\textwidth]{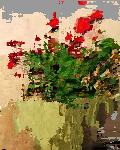}}
	\subfloat[\footnotesize{Proposal Flow.}]{
	\includegraphics[width=0.12\textwidth]{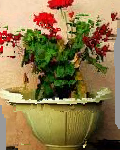}}
	\subfloat[\footnotesize{Source image.}]{
	\includegraphics[width=0.12\textwidth]{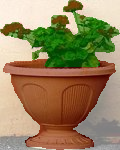}}
	\subfloat[\footnotesize{Target image.}]{
	\includegraphics[width=0.12\textwidth]{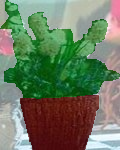}}
	\subfloat[\footnotesize{DSP.}]{
	\includegraphics[width=0.12\textwidth]{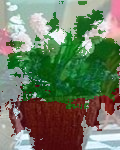}}
	\subfloat[\footnotesize{Proposal Flow.}]{
	\includegraphics[width=0.12\textwidth]{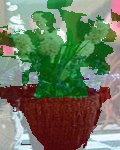}}
	\vfill

\caption{Examples of dense flow field on PASCAL parts. (a-b) Source images are warped to the target images using the dense correspondences estimated by (c) DSP~\cite{kim2013deformable} and (d) Proposal Flow w/ LOM, SS and HOG. (e-f) Similarly, annotated part segments for the source images are warped to the target images using the dense correspondences computed by (g) DSP and (h) Proposal Flow~(LOM w/ SS and HOG). \textbf{(Best viewed in color.)}} \vspace{-0.5cm}
\label{fig:voc_correspondence}	
\end{figure*}

\textbf{{Matching results on PASCAL parts.}}
We use the dataset provided by~\cite{zhou2015flowweb} where the images
are sampled from the PASCAL part dataset~\cite{chen2014detect}. Following~\cite{zhou2015flowweb}, we first measure part
matching accuracy using human-annotated part segments. For this
experiment, we measure the weighted IoU score between transferred
segments and ground truths, with weights determined by the pixel area
of each part (Table \ref{tb:voc}). To evaluate alignment accuracy, we
measure the PCK metric ($\alpha=0.05$) using keypoint annotations for
the 12 rigid PASCAL classes~\cite{xiang2014beyond}
(Table~\ref{tb:voc}). We use the same set of images as in the part
matching experiment. Proposal flow does better than existing approaches on images that
contain clutter~(e.g., background, instance-specific texture, occlusion), but in this dataset~\cite{zhou2015flowweb}, such elements are confined
to only a small portion of the images~(Fig.~\ref{fig:voc_correspondence}(a-b)), compared to the PF and the Caltech-101~\cite{fei2006one} datasets. This may be a reason that, for the PCK metric, our approach gives similar results to other methods. FlowWeb~\cite{zhou2015flowweb} gives better results than ours, but relies on a cyclic constraint across multiple images (at least, three images\footnote{FlowWeb~\cite{zhou2015flowweb} uses 100 images to find correspondences for one pair of images. That is, a single output of DSP~\cite{kim2013deformable} is refined using 9900 pairs of matches.}). FlowWeb uses the output of DSP~\cite{kim2013deformable} as initial correspondences, and refines them with the cyclic constraint. Since our method clearly outperforms DSP, using FlowWeb as a post processing would likely increase performance. Figure~\ref{fig:voc_correspondence} visualizes the part matching results.

 For more examples and qualitative results, see our project webpage: http://www.di.ens.fr/willow/research/proposalflow.

\begin{table*}[h!]
\centering
\footnotesize
\caption{PCK performance for a leave-one-out validation on the PF-WILLOW dataset.}
\vspace{-0.3cm}
\addtolength{\tabcolsep}{-2.0pt}
\begin{tabular}{l c c c c c c c c c c c }
\toprule
 Classes & car(S) & car(G) & car(M) & duc(S) & mot(S) & mot(G) & mot(M) & win(w/o C) & win(w/ C) & win(M) & Avg. \\
\midrule
\midrule
PCK & 0.95	&0.96	&0.99	&0.93	&0.88	&0.89	&0.91	&1.00	&1.00	&1.00	&0.95\\
\bottomrule
\end{tabular}
\label{tb:eva_gt_WILLOW}
\end{table*}

\begin{table*}[h!]
\centering
\footnotesize
\caption{PCK performance for a leave-one-out validation on the PF-PASCAL dataset.}
\vspace{-0.3cm}
\addtolength{\tabcolsep}{-3.0pt}
\begin{tabular}{l c c c c c c c c c c c c c c c c c c c c c}
\toprule
 Classes & aero & bike & bird & boat & bot & bus & car & cat & cha & cow & tab & dog & hor & mbik & pers & plnt & she & sofa & trai& tv & Avg.\\ 
\midrule
\midrule
PCK & 0.74	&0.89	&0.69	&0.91	&0.92	&0.90	&0.85	&0.83	&0.76	&0.81	&0.73	&0.75	&0.74	&0.75	&0.84	&0.83	&0.73	&0.83	&0.73	&0.86	&0.80\\
\bottomrule
\end{tabular}
\label{tb:eva_gt_PASCAL}
\end{table*}
\begin{figure}[h!]
\centering
	\includegraphics[width=0.45\textwidth]{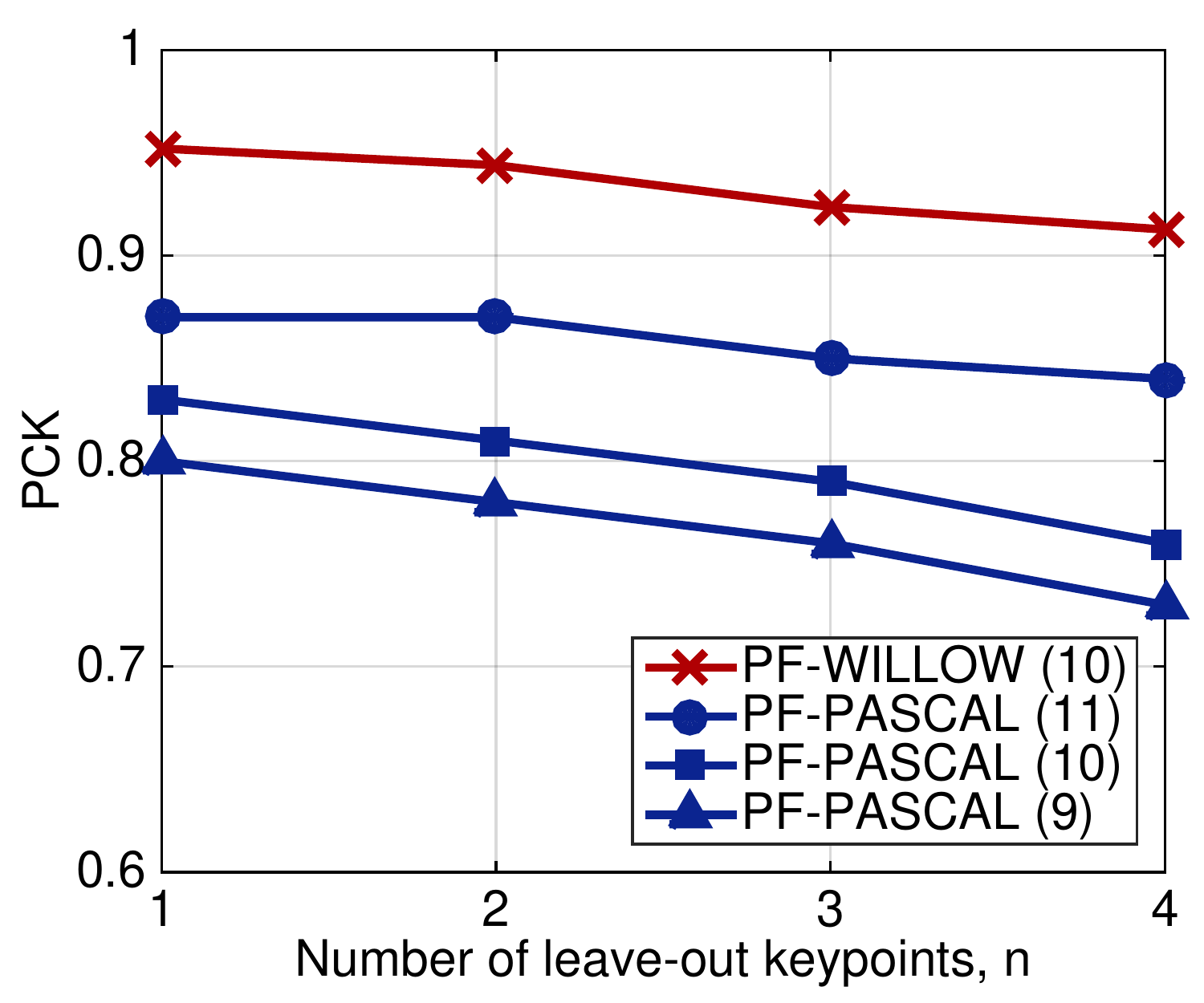}
\caption{Verification of ground-truth data using a leave-$n$-out validation. This shows the average PCK of 10 trials over all object classes. For this experiment, we leave out $n$ randomly selected keypoints per each pair, and then measure PCK scores between the estimated correspondences (using TPS warps) of the leave out keypoints and their ground-truth annotations. \textbf{(Best viewed in color.)}}
\label{fig:eva_gt}	
\end{figure}

\subsection{Quality of generated ground-truth correspondence}\label{subsec:quality}
Of course, our ``ground truth" for the PF datasets is only approximate, since it is obtained by interpolation. We evaluate its quality using a leave-$n$-out validation: When generating ground-truth dense correspondences using TPS warping as in Section~\ref{subsec:gt}, we leave out $n$ randomly selected keypoints per each pair (e.g., $n$ among 10 keypoints in the PF-WILLOW dataset), and then evaluate PCK ($\alpha=0.1$) between the approximated correspondences~(using TPS warps)~of the leave-out keypoints and their ground-truth annotations. The average PCK of 10 trials over all object classes is shown in Fig.~\ref{fig:eva_gt}. The number in parentheses denotes the number of ground-truth keypoints. For the PF-PASCAL, each image pair has a different number of keypoints. We see that using more keypoint annotations improves the quality of generated ground truth. Note that perfect score would be 1.0. In Tables~\ref{tb:eva_gt_WILLOW} and \ref{tb:eva_gt_PASCAL}, we show the PCK results for a leave-\emph{one}-out validation. The average PCK scores are $0.95$ and $0.80$ on the PF-WILLOW and PF-PASCAL, respectively. These numbers are quite reasonable, and validates the use of our ground-truth data using TPS. 

\subsection{Object proposals vs.~sliding windows}
Our experiments show that proposal flow outperforms state-of-the-art methods such as SIFT flow~\cite{liu2011sift}, DSP~\cite{kim2013deformable}, and DeepFlow~\cite{weinzaepfel2013deepflow}: Note that these methods all employ a sort of sliding window strategies for matching (i.e., regular  sampling with a fixed stride, and in particular, DeepFlow~\cite{weinzaepfel2013deepflow} with stride 1). Figures~\ref{fig:benchmark}(a) and~\ref{fig:benchmark-auc} evaluate SW within our approach, where we make proposals by placing windows on a regular grid across predefined 5 scales and 5 aspect ratios with a uniform stride (following~\cite{hosang2015what}). The PCR and mIoU@$k$ plots show that object proposals clearly outperform SW with the same number of regions. In Table~\ref{tb:caltech}, we can see that 1)~the proposal flow method with SW already outperforms competing algorithms, and 2)~it further benefits from the use of SS to go from 0.47 to 0.50 in terms of the IoU metric. Note that this metric focuses on the foreground matching quality~\cite{kim2013deformable}, implying that the use of object proposals helps in matching foreground regions. The advantage can be clearly seen with more cluttered images. For example, LOM with SW and SS on the PF-WILLOW gives PCK ($\alpha=0.1$) of 0.42 and 0.56, respectively, as shown in Table~\ref{tb:pck}. The superior performance comes from the effective use of geometric contextual information as well as that of object proposals.

\section{Discussion}
We have presented a robust region-based semantic flow method, called proposal
flow, and shown that it can effectively be mapped to pixel-wise
dense correspondences. We have also introduced the PF datasets for semantic flow, and shown that they provide a reasonable benchmark for a semantic
flow evaluation without extremely expensive manual annotation of full ground truth. 
Our benchmarks can be used to evaluate region-based semantic flow methods and also pixel-based ones, and experiments with the PF datasets  
demonstrate that proposal flow substantially outperforms
existing semantic flow methods. Experiments
with Caltech-101, the PASCAL parts, and Taniai's datasets further validate these results.

\ifCLASSOPTIONcompsoc
  \section*{Acknowledgments}
\else
  \section*{Acknowledgment}
\fi

This work was supported by ERC grants VideoWorld and Allegro, and the Institut Universitaire de France.

\ifCLASSOPTIONcaptionsoff
  \newpage
\fi

\bibliographystyle{IEEEtran}
\bibliography{proposal_flow}

\begin{IEEEbiographynophoto}{Bumsub Ham}
is an an Assistant Professor of Electrical and Electronic Engineering at Yonsei University in Seoul, Korea. He received the B.S. and Ph.D. degrees in Electrical and Electronic Engineering from Yonsei University in 2008 and 2013, respectively. From 2014 to 2016, he was Post-Doctoral Research Fellow with Willow Team of INRIA Rocquencourt, {\'E}cole Normale Sup{\'e}rieure de Paris, and Centre National de la Recherche Scientifique. His research interests include computer vision, computational photography, and machine learning, in particular, regularization and  matching, both in theory and applications.
\end{IEEEbiographynophoto}
\begin{IEEEbiographynophoto}{Minsu Cho}
is an Assistant Professor of computer science and engineering at POSTECH in Pohang, South Korea. He obtained his PhD degree in Electrical Engineering and Computer Science from Seoul National University in 2012. Before joining POSTECH in 2016, he worked as an Inria starting researcher in the ENS/Inria/CNRS Project team WILLOW at École Normale Superiéure, Paris, France. His research lies in the areas of computer vision and machine learning, especially in the problems of object discovery, weakly-supervised learning, and graph matching.  
\end{IEEEbiographynophoto}
\begin{IEEEbiographynophoto}{Cordelia Schmid}
holds a M.S. degree in Computer Science from the
University of Karlsruhe and a Doctorate, also in Computer Science,
from the Institut National Polytechnique de Grenoble (INPG). Her
doctoral thesis received the best thesis award from INPG in 1996. 
Dr. Schmid was a post-doctoral research assistant in the Robotics
Research Group of Oxford University in 1996--1997. Since 1997 she has
held a permanent research position at INRIA Grenoble Rhone-Alpes,
where she is a research director and directs an INRIA team. Dr. Schmid
is the author of over a hundred technical publications. She has been
an Associate Editor for IEEE PAMI (2001--2005) and for IJCV
(2004--2012), editor-in-chief for IJCV (2013---), a program chair of
IEEE CVPR 2005 and ECCV 2012 as well as a general chair of IEEE CVPR
2015 and ECCV 2020. In 2006, 2014 and 2016, she was awarded 
the Longuet-Higgins prize for fundamental contributions in computer
vision that have withstood the test of time. She is a fellow of
IEEE. She was awarded an ERC advanced grant in 2013, the Humbolt
research award in 2015 and the Inria \& French Academy of science
Grand Prix in 2016.
\end{IEEEbiographynophoto}
\begin{IEEEbiographynophoto}{Jean Ponce}
is a Professor at {\'E}cole Normale Sup{\'e}rieure (ENS) and PSL Research University in Paris, France, where he leads a joint ENS/INRIA/CNRS research team, WILLOW, that focuses on computer vision and machine learning. Prior to this, he served for over 15 years on the faculty of the Department of Computer Science and the Beckman Institute at the University of Illinois at Urbana-Champaign. Dr. Ponce is the author of over 150 technical publications, including the textbook ``Computer Vision: A Modern Approach", in collaboration with David Forsyth. He is a member of the Editorial Boards of Foundations and Trends in Computer Graphics and Vision, the International Journal of Computer Vision, and the SIAM Journal on Imaging Sciences. He was also editor-in-chief of the International Journal on Computer Vision (2003-2008), an Associate Editor of the IEEE Transactions on Robotics and Automation (1996-2001), and an Area Editor of Computer Vision and Image Understanding (1994-2000). Dr. Ponce was Program Chair of the 1997 IEEE Conference on Computer Vision and Pattern Recognition and served as General Chair of the year 2000 edition of this conference. In 2003, he was named an IEEE Fellow for his contributions to Computer Vision, and he received a US patent for the development of a robotic parts feeder. In 2008, he served as General Chair for the European Conference on Computer Vision.

\end{IEEEbiographynophoto}

\end{document}